\documentclass[10pt,letterpaper]{article}

\usepackage{fullpage}
\usepackage{cite}
\usepackage{algorithmic}
\usepackage{textcomp}

\usepackage{amsmath, graphicx, epsfig, color, amsfonts, algorithm, amssymb, mathtools, array, url, multirow, booktabs}
\usepackage{soul}
\usepackage{amsthm}
\usepackage{lineno}

\usepackage[font=footnotesize]{caption}
\usepackage{wrapfig}

\usepackage{subcaption}
\usepackage{float}

\usepackage[normalem]{ulem}

\usepackage[inline]{enumitem}

\usepackage{version,xspace}
\usepackage{tikz}
\usetikzlibrary{arrows.meta, positioning, fit, backgrounds, shapes.geometric}

\def\expt{\mathbb{E}}

\newcommand{\subscr}[2]{#1_{\textup{#2}}}
\newcommand{\supscr}[2]{#1^{\textup{#2}}}

\def \bs {\boldsymbol}
\def \mc {\mathcal}
\def\tran{^\top}

\def \mb {\mathbb}

\def\expt{\mathbb{E}}

\usepackage{epstopdf}

\newcommand{\rev}[1]{\textcolor{black}{#1}}

\begin{document}
\title{Skill-informed Data-driven Haptic Nudges for High-dimensional Human Motor Learning}
\author{
Ankur Kamboj$^1$, Rajiv Ranganathan$^2$, Xiaobo Tan$^1$, and Vaibhav Srivastava$^1$
\thanks{This work was supported by NSF Grant CMMI-1940950.\\
$^{1}$A. Kamboj ({\tt\small ankurank@msu.edu}), X. Tan ({\tt\small xbtan@msu.edu}), and V. Srivastava ({\tt\small vaibhav@msu.edu}) are with the Department of Electrical and Computer Engineering, Michigan State University, USA.\\
$^2$R. Ranganathan ({\tt\small rrangana@msu.edu}) is with the Department of Kinesiology and the Department of Mechanical Engineering, Michigan State University, USA.}
}
\date{}
\maketitle

\begin{abstract}
\rev{In this work, we propose a data-driven framework to design optimal haptic nudge feedback leveraging the learner's estimated skill to address the challenge of learning a novel motor task in a high-dimensional, redundant motor space. A nudge is a series of vibrotactile feedback delivered to the learner to encourage motor movements that aid in task completion.}
We first model the stochastic dynamics of human motor learning \rev{under haptic nudges} using an Input-Output Hidden Markov Model (IOHMM), which explicitly decouples latent skill evolution from observable performance measures. Leveraging this predictive model, we formulate the haptic nudge feedback design problem as a Partially Observable Markov Decision Process (POMDP). This allows us to derive an optimal nudging policy that minimizes long-term performance cost \rev{and implicitly guides the learner toward superior skill states}. \rev{We validate our approach through a human participant study ($N=30$) involving a high-dimensional motor task rendered through a hand exoskeleton.} Results demonstrate that participants trained with the POMDP-derived policy exhibit significantly accelerated \rev{movement efficiency and endpoint accuracy} compared to groups receiving heuristic-based feedback or no feedback. Furthermore, synergy analysis reveals that the POMDP group discovers efficient low-dimensional motor representations more rapidly.
\end{abstract}

\section{Introduction}
\rev{The integration of robotic systems into human environments presents a transformative opportunity to support skill acquisition and motor learning through automated and personalized physical assistance. However, facilitating learning in high-dimensional \emph{de-novo} (novel) tasks, such as manipulating a complex tool or recovering motor function post neurological injury, poses a formidable computational and cognitive challenge~\cite{wolpert2011principles}. De-novo learning necessitates the discovery of novel movement strategies, which fundamentally distinguishes it from the simple reduction of kinematic task errors observed in low-dimensional reaching tasks~\cite{haith2013theoretical, krakauer2011human}. Furthermore, the human motor system exhibits massive kinematic redundancy, meaning the solution space for any high degrees of freedom (DoF) task is vast~\cite{ranganathan2013learning}. Consequently, learners must navigate this highly complex manifold to identify optimal motor coordination strategies. Designing automated guidance for this learning process is an open challenge since determining the optimal movement for such tasks is non-trivial.}

\rev{To assist in this complex skill acquisition process, haptic feedback has proven to be a powerful interventional modality that guides human movements without fostering dependence on robotic assistance~\cite{marchal2009review, williams2014motor}.
Haptic guidance designs for physical human robot interaction regimes can be broadly categorized into two approaches: (i) expert-in-the-loop, and (ii) task-performance-based automated approaches. The first category involves field experts and professionals designing haptic guidance systems informed by their intuition, experience, and knowledge~\cite{turolla2013haptic, lin2016development, esmatloo2020fingertip}. While effective for constrained movements, this ``top-down" approach scales poorly to high-DoF tasks where finding optimal motions to instruct precise motor coordination strategies is overwhelmingly complex due to the aforementioned redundancy.} The second category exploits directly observable task performance metrics like task errors, movement time, smoothness, or task success to design haptic robotic feedback~\cite{basalp2021haptic, sigrist2013augmented, sullivan2021haptic, rowe2017robotic}. Although this reactive haptic feedback strategy is intuitive and easier to automate, it can often be misleading as performance is a noisy measure of true skill, especially for high DoF tasks. This assistance might overfit to the user, helping them perform well temporarily (low errors) by using a bad, compensatory habit (low skill). The simplistic task performance might not capture important features of the underlying latent skill state, thus leading learners toward suboptimal solutions. This prevents mastery of high-dimensional skill and ultimately hinders long-term skill development.

To address these limitations, we propose automated haptic nudge assistance designs informed by latent skill rather than task performance corrections. \rev{In our work, the robotic nudge feedback is a series of two vibrations delivered to a finger during the early phase of a trial, suggesting that the participant should move that finger more to accomplish the task. We specifically employ discrete vibrotactile nudges over continuous force feedback to maintain active motor exploration and mitigate the risk of learner dependency. While continuous force feedback effectively guides users along reference trajectories, it fundamentally alters the dynamics of the task. This physical intervention often induces the ``guidance hypothesis"~\cite{lee1990role}, where learners rely on robotic assistance and fail to internalize the task dynamics, resulting in poor long-term retention~\cite{marchal2009review, williams2014motor}. Conversely, discrete haptic feedback acts as an informational cue without the physical constraints. Prior research demonstrates that vibrotactile feedback effectively conveys proprioceptive information under cognitive load~\cite{vargas2021static}, drives targeted biomechanical adaptations in dynamic gait retraining~\cite{escamilla2021exploration}, and prompts voluntary limb movement in stroke rehabilitation~\cite{signal2023haptic}. By relying on discrete nudges, our framework requires learners to actively decode the haptic stimulus and execute an intrinsic motor plan, ensuring independent exploration of the redundant task manifold.
} 

Central to our approach is the recognition that motor learning is a latent process, in which observable performance is a noisy emission of the underlying, evolving motor learning and control strategy~\cite{tenison2016modeling}. Drawing on human skill modeling literature~\cite{yang1994hidden, tenison2016modeling, zhang2020multilevel}, we formalize motor learning as a progression through discrete latent states~\cite{yuh2024classification}, paralleling the cognitive, associative, and autonomous phases proposed by Fitts and Posner~\cite{fitts1967human}.
While standard Hidden Markov Models (HMM) have successfully tracked such transitions in skill acquisition~\cite{tenison2016modeling, megali2006modelling, french2024determining, ghonasgi2021capturing} and neural motor planning~\cite{kemere2008detecting, kirchherr2023bayesian}, effective robotic nudge feedback requires understanding how external interventions influence these transitions. Thus, we employ a probabilistic predictive model of human motor learning using an Input-Output Hidden Markov Model (IOHMM). This structure allows us to causally link specific haptic nudges (inputs) to probabilities of skill improvement, effectively separating execution noise from true latent capability.

Several mechanistic models have been proposed in the literature to explain the stochastic human motor learning process~\cite{smith2006interacting, pierella2019dynamics, kamboj2024human}. One salient feature of these models is the Markovian evolution of latent states as a function of certain inputs to the model. These models can be abstracted into a probabilistic finite-state machine defined over a quantized state space with event-triggered temporal sampling. The IOHMM operates on this abstraction and incorporates the resulting finite-state machine as its latent dynamics.
This probabilistic framework allows us to treat the haptic tutoring as a sequential decision-making problem under uncertainty. Leveraging the IOHMM, we formulate a Partially Observable Markov Decision Process (POMDP) to derive an optimal nudging policy. Similar to prior work that utilized POMDPs to infer latent user fatigue during rehabilitation~\cite{huq2011decision}, our formulation models the evolution of latent motor skill over the duration of the motor task being performed by the learner. The objective is to prescribe a sequence of haptic feedback nudges that minimizes long-term performance cost, and implicitly guides the learner toward effective regions of the latent skill space. This enables a principled, data-driven approach, going beyond performance or expert-designed heuristics, for personalizing haptic feedback design in novel skill acquisition paradigms.

Building on the preliminary findings of our POMDP-based optimal nudge feedback~\cite{kamboj2026toward}, this work investigates the potential of the skill-informed haptic feedback design in accelerating skill acquisition and task performance for high-dimensional, de-novo motor tasks. We distinguish this study from previous work~\cite{kamboj2026toward} through the following advancements:
\begin{enumerate*}[label=(\roman*)]
    \item we introduce a new participant group training on a heuristic nudging policy to benchmark performance against no-feedback and POMDP-based optimal nudge feedback, and increase the sample size to better delineate the group effects,
    \item we employ additional analysis to evaluate how effectively different group participants are discovering the task space manifold,
    \item we analyzed the evolution of participants' estimated latent skill states over the experiment trials,
    \item we analyze the participants' motor output performance using a reaching error metric across the experiment trials, and
    \item we expand upon the IOHMM employed to capture the human motor learning behavior, the training algorithm used, and provide interpretability to the estimated model parameters.
\end{enumerate*}
In summary, our contributions are threefold as follows.
\begin{enumerate}
    \item We introduce an IOHMM framework that captures the dynamics of high-dimensional motor learning with motor skill as a latent state,  haptic nudge, and task type as inputs, and performance measures as noisy outputs.
    \item We formulate a POMDP to compute an optimal, skill-aware haptic nudging policy, providing a principled alternative to heuristic feedback designs.
    \item We validate our approach through a human participant study, demonstrating that our POMDP-derived policy significantly expedites motor skill learning and enhances \rev{both endpoint accuracy and movement efficiency} compared to both no-feedback and a heuristic feedback strategy.
\end{enumerate}
The rest of the paper is organized as follows. Section~\ref{sec:background} details the experimental setup and theoretical preliminaries. Section~\ref{sec:IOHMM} describes the behavioral model of human motor learning as an IOHMM in detail. Then, in Section~\ref{sec:pomdp_design}, we detail the derived POMDP and design the performance heuristics-based and optimal nudging policy. Section~\ref{sec:results} presents our experimental findings, followed by a discussion of the results and concluding remarks in Section~\ref{sec:discussion}.
\begin{figure*}[th!]
    \centering
    \begin{subfigure}{0.25\textwidth}
	    \centering
        \includegraphics[width=0.7\linewidth, height=1\linewidth, keepaspectratio]{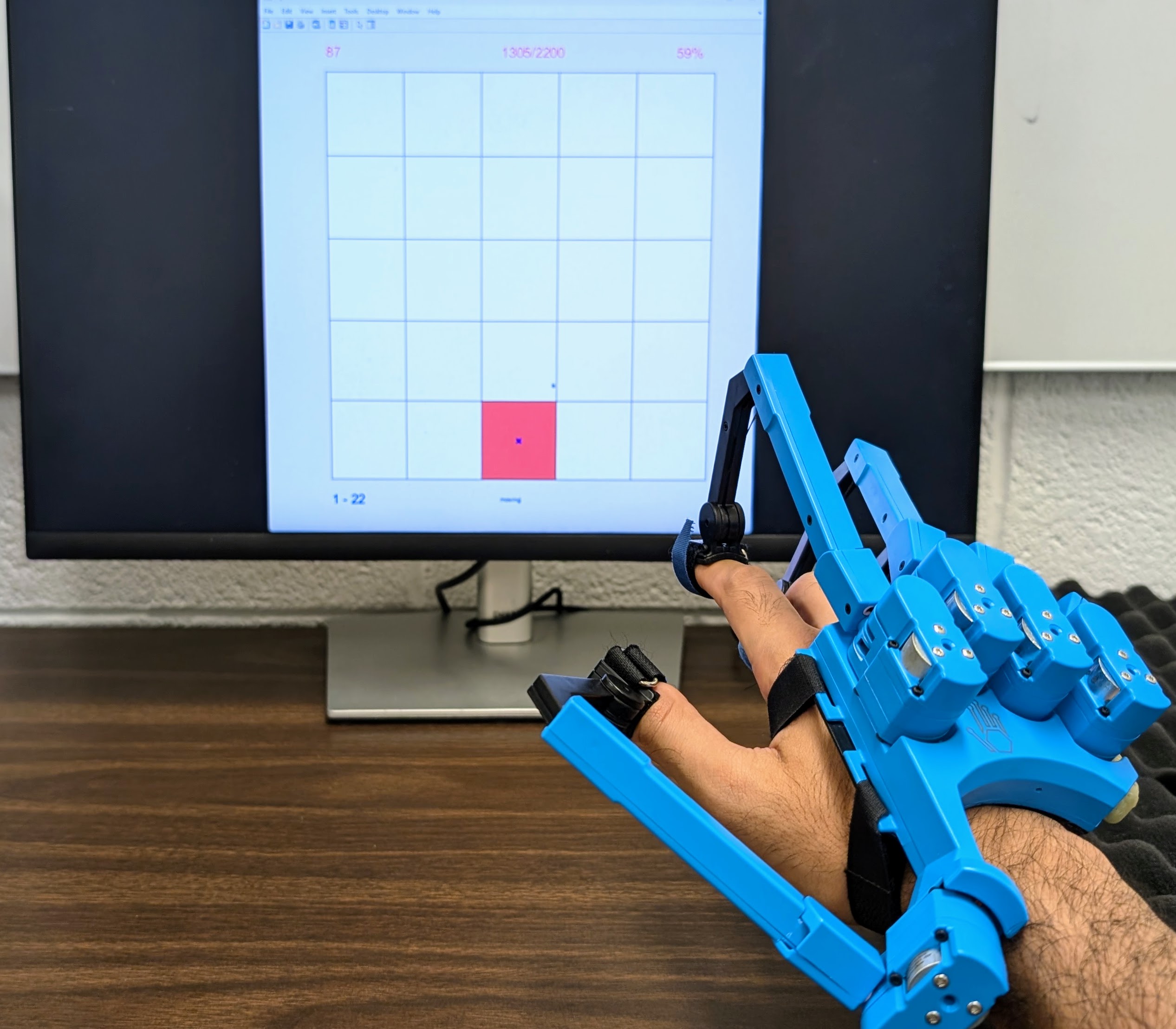}   
        \vspace{1.3em}     
        \subcaption{}
    \end{subfigure}\hspace{-3.0em}%
    ~
    \begin{subfigure}{0.25\textwidth}       
	    \centering
        \includegraphics[width=1\linewidth, height=1\linewidth, keepaspectratio]{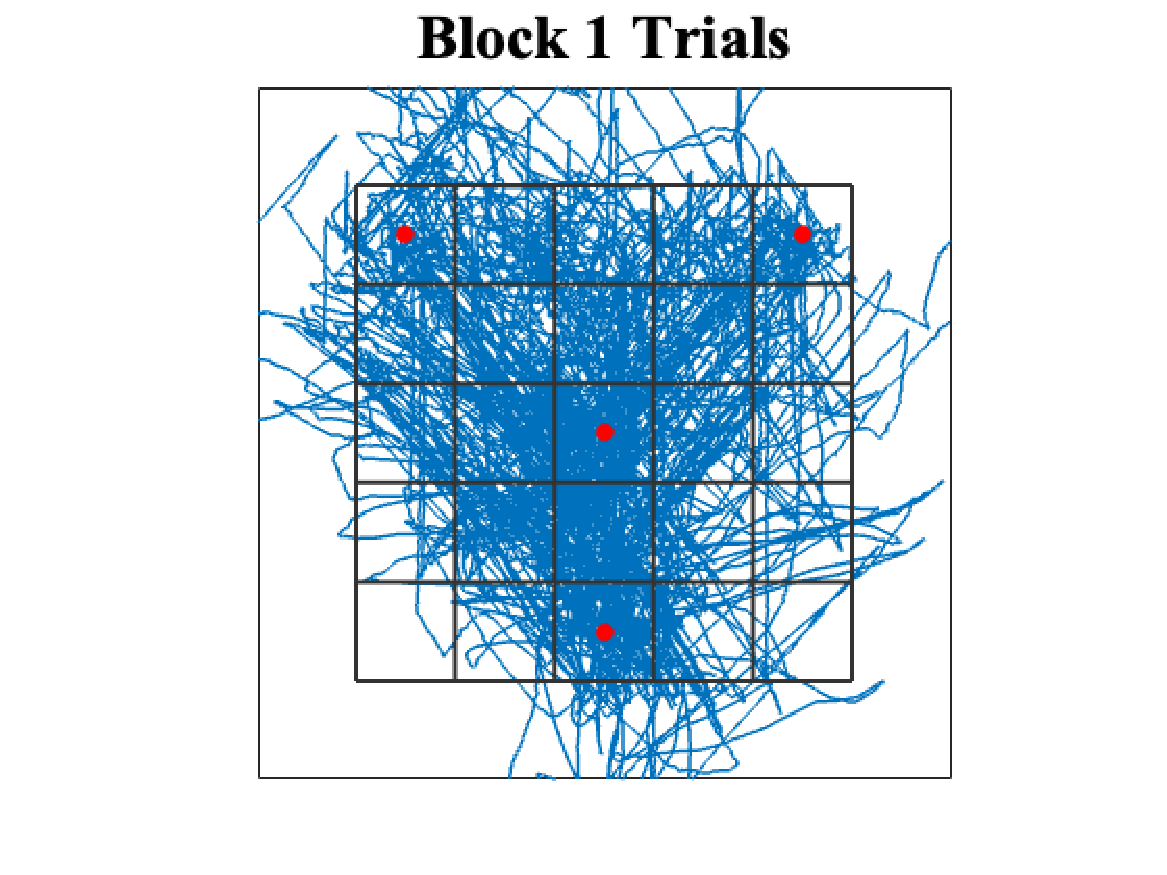}        
        \subcaption{}
    \end{subfigure}\hspace{-2.8em}%
    ~
    \begin{subfigure}{0.25\linewidth}
	    \centering
        \includegraphics[width=1\linewidth, height=1\linewidth, keepaspectratio]{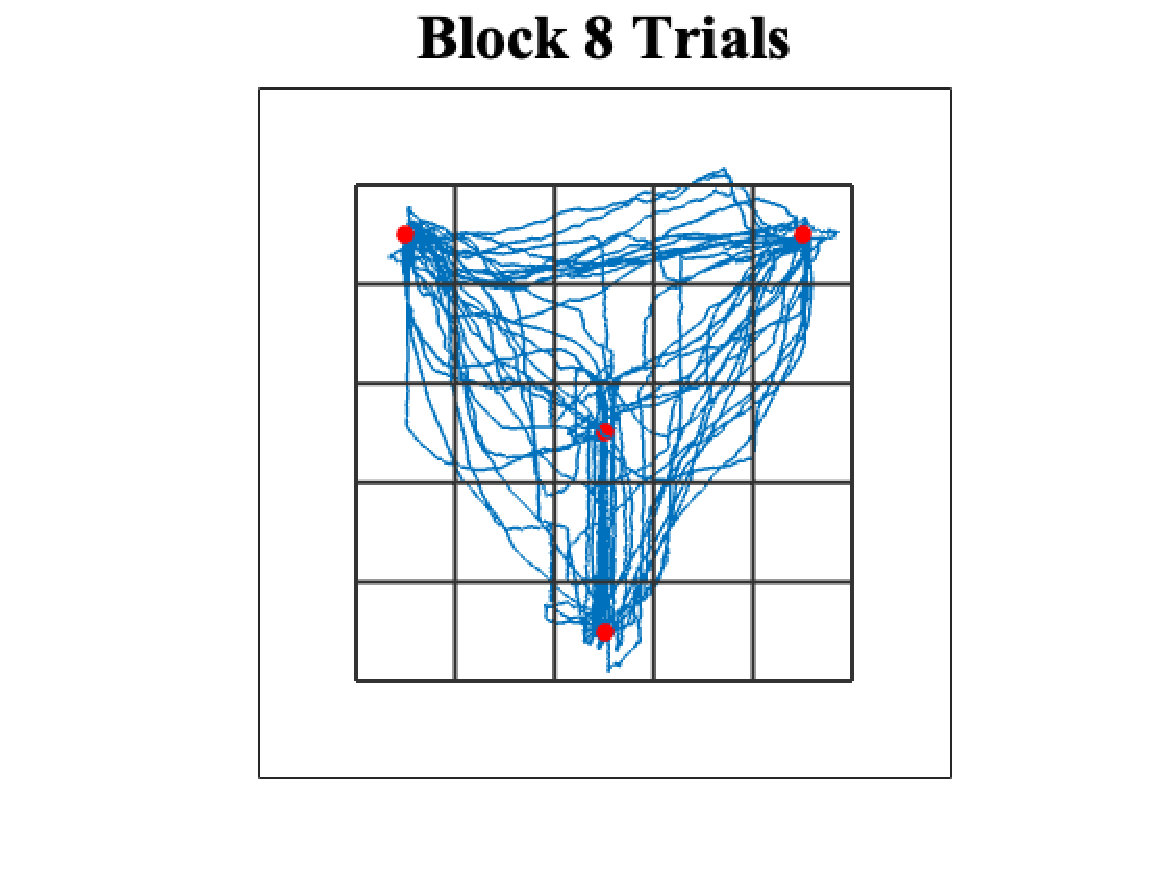}        
        \subcaption{}
    \end{subfigure}\hspace{-2.8em}%
    ~
    \begin{subfigure}{0.25\linewidth}
	    \centering
        \includegraphics[width=1\linewidth, height=1\linewidth, keepaspectratio]{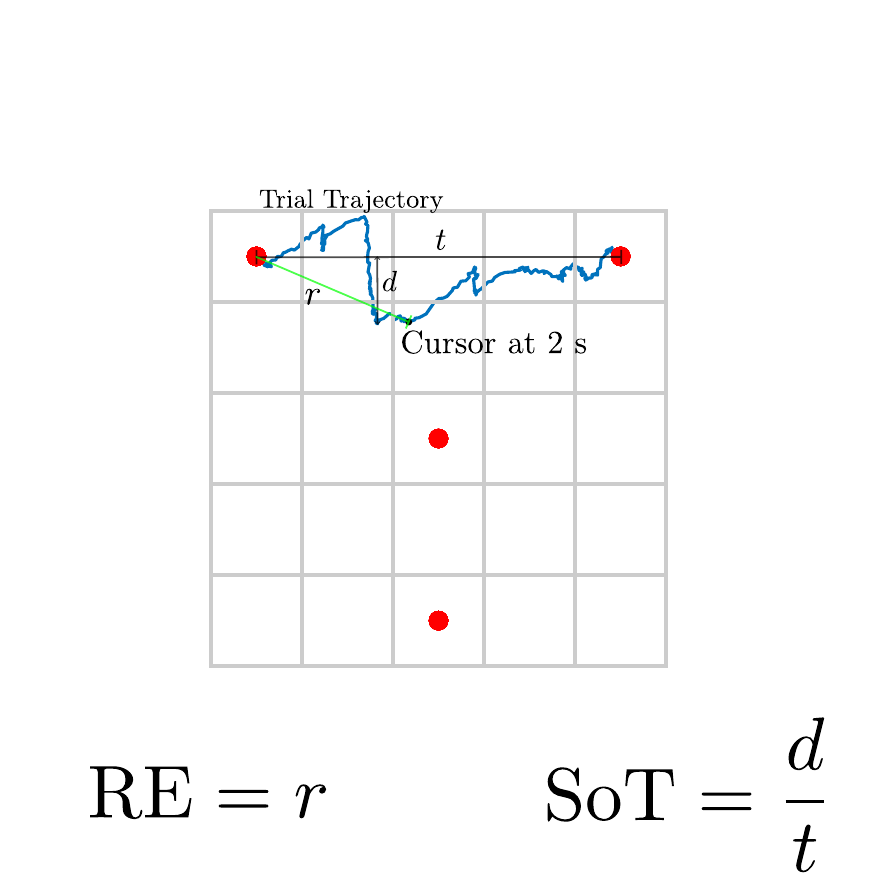}    
        \vspace{-2.5em}       
        \subcaption{}
        \label{fig:RESOT_illus}
    \end{subfigure}
    \caption{\textbf{Target Capture Task:} (a) shows a participant performing our target capture task with the SenseGlove DK1 exoskeleton strapped to their right hand. (b) and (c) show the cursor trajectories during the first and eighth blocks of the game with the red dots representing the target locations. The trajectories get straighter as the participant learns to control the cursor by the end of the experiment session. (d) shows how the performance metrics, Reaching Error and Straightness of Trajectory, are computed.}
    \label{fig:target_capture_game}
\end{figure*}
\section{Background}    \label{sec:background}
This section details the experimental framework and the performance metrics used to evaluate human performance during the acquisition of a novel, high-dimensional motor skill. 
Subsequently, we outline the stochastic modeling approach adopted in this paper.

\subsection{Experimental Setup}     \label{sec:target_capture_game}
We adopt a Body-Machine Interface (BoMI) paradigm similar to those described in~\cite{mosier2005remapping, ranganathan2013learning}. 
Participants wore a SenseGlove DK1 (manufactured by SenseGlove, a Netherlands-based company), a non-invasive hand exoskeleton capable of tracking $20$ distinct degrees of freedom (DoF) across the hand joints. Additionally, the device can deliver independent vibrotactile haptic feedback to each of the five fingertips. This high-dimensional joint space is then projected into a lower-dimensional space, represented by a cursor on a screen. This projection is governed by the linear mapping
\begin{equation}
    \dot{\bs x} = C \bs u,
\end{equation}
where $\dot{\bs x} \in \mathbb{R}^n$ represents the velocity of the screen cursor, $\bs u = \dot{\bs q} \in \mathbb{R}^m$ denotes the finger joint velocities derived from the joint angles $\bs q$, and $C \in \mathbb{R}^{n \times m}$ serves as the BoMI mapping matrix.
\rev{Participants are instructed to perform a \emph{target capture task}\footnote{A video footage of the target capture task is available at \url{https://youtu.be/WIDRwqpE9Sk}} (see Fig.~\ref{fig:target_capture_game}), in which they must coordinate their finger joint movements to navigate the cursor toward sequentially appearing targets.} Because the mapping from the $20$-dimensional hand space to the $2$-dimensional cursor space is initially unknown to the user, the task enforces a high-dimensional learning process. Success requires the participant to discover and internalize specific coordinated hand movements consistent with the mapping matrix $C$. \rev{A haptic nudge, if prescribed to the participant, consists of a tactile stimulus: two sequential vibration bursts, each lasting $150~$ms, separated by $2000~$ms, delivered to the fingertip prescribed by the active feedback policy. The strength of this nudge is chosen such that it is clearly perceived by the participant, but doesn't impact the finger motion.}
The experimental protocol is divided into three distinct phases:
\begin{itemize}
    \item \textbf{Calibration:} To customize the mapping, participants mimic a series of American Sign Language (ASL) alphabet hand postures. We apply Principal Component Analysis (PCA) to the recorded hand posture data and construct the mapping matrix $C$ using the first and the third principal components. The coordinate system is centered on the task space, represented as a $5\times 5$ unit game window, such that the user's mean hand posture corresponds to the center of the game screen. One unit is the width of a grid square, computed to accommodate $1$-standard deviation of the calibration data points and to ensure all potential targets are comfortably within the user's reach.
    
    \item \textbf{Familiarization:} A brief acclimation period of approximately $6$ seconds is provided. During this time, participants are free to move the cursor within the task space (Fig.~\ref{fig:target_capture_game}). This duration is intentionally short to allow users to get a feel for the system dynamics without inducing significant motor learning or adaptation effects.
    
    \item \textbf{Training:} The primary gameplay phase comprises $8$ blocks, each containing $60$ individual trials. A trial consists of a single reaching movement toward a prescribed target; upon successful capture, a new target is immediately presented. A capture is registered if the cursor remains stable within the target's grid square—defined as position changes of less than $0.0025$ units over $15$ consecutive samples at $100$ Hz. The targets are prescribed uniformly randomly from a set of $4$ targets $\mc T = \{\mc T_1, \mc T_2, \mc T_3, \mc T_4\}$ at $\{(0.5,4.5), (2.5,2.5), (2.5,0.5), (4.5,4.5)\})$ units of the task space grid, respectively.
\end{itemize}

\subsection{Performance Metrics}
\rev{To comprehensively evaluate the motor learning process, we utilize endpoint accuracy, which measures task success, and movement efficiency, which characterizes the optimality of the underlying motor control strategy. In highly redundant tasks, learners frequently achieve high accuracy using suboptimal, compensatory movements. Therefore, evaluating both metrics is essential to verify true skill acquisition. Consequently, we quantify task performance using two primary metrics~\cite{kamboj2024human}: Reaching Error (\texttt{RE}) to evaluate endpoint accuracy, and Straightness of Trajectory (\texttt{SoT}) to evaluate movement efficiency (Fig.~\ref{fig:RESOT_illus})}. 
\begin{itemize}
    \item \texttt{RE} is defined as the Euclidean distance between the cursor and the target center at the end of a trial. The end of the trial is defined as the instant when the cursor position did not change by more than $0.0025$ units for $15$ consecutive samples, or $2$ seconds after the start of the movement, whichever occurs first.
    \item \texttt{SoT} assesses the efficiency of the path. It is calculated as the aspect ratio between the maximum perpendicular deviation of the trajectory from the ideal straight line (connecting the start and end points) and the length of that straight line.
\end{itemize}

\subsection{IOHMM and POMDP Framework}
To model the stochastic dynamics of the user's internal skill-state, we employ an Input-Output Hidden Markov Model (IOHMM)~\cite{bengio1996input}. This probabilistic graphical model describes the evolution of a discrete latent state sequence $\{s_k \in \mathcal{S}\}_{k \in \mathbb{N}}$, where $\mc S$ is the state space, and the transition to a future state $s_{k+1}$ depends on both the current state $s_{k}$ and an exogenous control input or action $a_k \in \mathcal{A}$, where $\mc A$ is the action space. Crucially, the true system state is inaccessible for direct measurement; it is hidden. Instead, inference is performed via observations $\{o_k \in \Omega\}_{k\in \mathbb{N}}$, where $\Omega$ is the observation space. These are modeled through the conditional probability distribution $\mathbb{P}(o|s, a)$, representing the likelihood of observing output $o$ given the system is in state $s$ and receives input $a$. We assume $\mathcal{S}$ and $\mathcal{A}$ are finite sets.

Given a historical dataset of input-output pairs $\{a_k, o_k\}_{k\in \mathbb{N}}$, the IOHMM is trained to learn the state transition dynamics $T(s, a, s') = \mathbb{P}(s'|s, a)$, the observation probability matrix $O(s, a, o) = \mathbb{P}(o|s, a)$, and the initial state distribution. This parameter estimation is typically achieved via the extended Baum-Welch algorithm~\cite{bengio1996input}, which adapts the standard Expectation-Maximization (EM) algorithm to account for the input-dependent structure of the IOHMM.

To transition from system modeling to active control, we extend the IOHMM into a Partially Observable Markov Decision Process (POMDP). The POMDP framework introduces an agent that optimizes the selection of inputs based on a defined objective. Formally, the POMDP is characterized by the tuple $\langle \mathcal{S}, \mathcal{A}, T, R, \Omega, O\rangle$. In this formulation, $\mathcal{S}, \mathcal{A}, T, \Omega,$ and $O$ retain their definitions from the IOHMM, while $R(s,a)$ represents the reward function. This function quantifies the agent's performance goals, assigning positive scalar values to beneficial state-action pairs and penalties (costs) to undesirable outcomes, thereby guiding the optimal control policy.

\rev{The IOHMM-POMDP framework offers specific analytical and computational advantages for modeling and influencing the human motor learning process. Unlike deep reinforcement learning or recurrent neural networks for sequence modeling and control, IOHMM enables sample-efficient parameter estimation from limited human data while preserving interpretability. The state transitions and emission models allow us to explicitly quantify the effects of targeted haptic interventions (Section~\ref{sec:results}). Furthermore, unlike standard Markov Decision Processes (MDPs) that assume fully observable states, the POMDP formally accounts for execution noise. By decoupling noisy task performance metrics from discrete latent skill states, the derived policy optimizes for long-term motor adaptation rather than over-correcting for transient kinematic errors.}

\section{Human Motor Learning Behavioral Model} \label{sec:IOHMM}
To develop a behavioral model of the human learners' response to haptic nudging, we employ a dynamic model of motor performance using an IOHMM.
We detail the IOHMM model employed in this work, including its structure and the associated parameters. This is followed by the training/fitting techniques used and the interpretation of this model estimated from the experiment data.

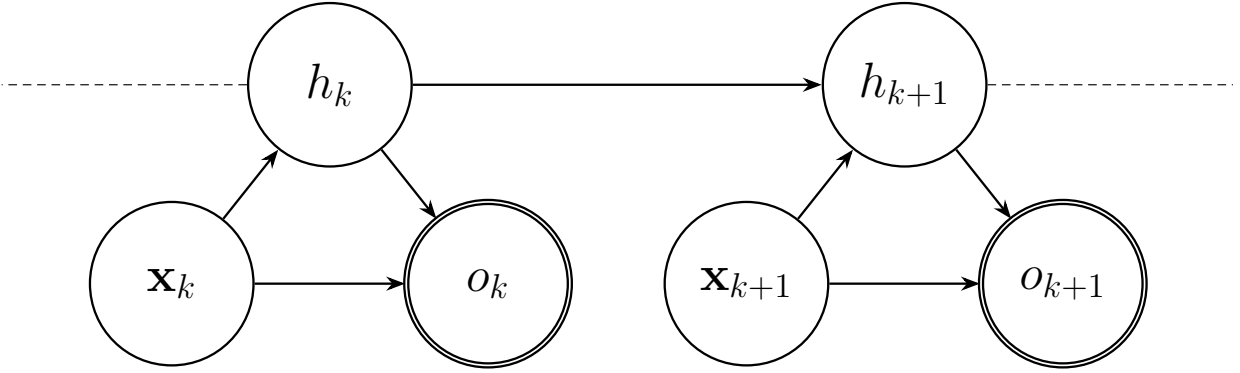
\begin{figure}[h!]
    \centering
    \resizebox{\columnwidth}{!}{%
    \begin{tikzpicture}[
        font=\LARGE,
        node distance=1.2cm and 2cm,
        latent/.style={ellipse, draw, thick, minimum size=2cm},
        observable/.style={ellipse, draw, double, thick, minimum size=2cm, inner sep=2pt},
        input/.style={ellipse, draw, thick, minimum size=2cm},
        every edge/.style={draw, ->, >=Stealth, thick}
    ]

    \node[latent] (h_k) {$h_k$};
    \node[observable, below right=1cm and 0.5cm of h_k]  (o_k) {$o_{k}$};
    \node[input, below left=1cm and 0.5cm of h_k]  (u_k) {$\mathbf{x}_{k}$};

    \node[latent, right=5cm of h_k] (h_k+1) {$h_{k+1}$};
    \node[observable, below right=1cm and 0.5cm of h_k+1]  (o_k+1) {$o_{k+1}$};
    \node[input, below left=1cm and 0.5cm of h_k+1]  (u_k+1) {$\mathbf{x}_{k+1}$};

    \path (h_k) edge (o_k);
    \path (u_k) edge (h_k);
    \path (u_k) edge (o_k);

    \path (h_k) edge (h_k+1);

    \path (h_k+1) edge (o_k+1);
    \path (u_k+1) edge (h_k+1);
    \path (u_k+1) edge (o_k+1);

    \draw[densely dashed] (h_k.west) -- ++(-3cm, 0);
    \draw[densely dashed] (h_k+1.east) -- ++(3cm, 0);

    \end{tikzpicture}
    }
    \caption{\textbf{Simplified IOHMM:} Human motor learning behavior modeled as an IOHMM shows how the hidden motor skill state $h_k$ transitions to $h_{k+1}$ under the effect of the input $\bf{x}_k = \{u_k: \texttt{slope}_k, a_k: \texttt{nudge}_k\}$, leading to the emissions $o_k = \{\texttt{RE}_k, \texttt{SoT}_k\}$ under the influence of the same input $\bf{x}_k$.}
    \label{fig:iohmm}
\end{figure}
\subsection{Behavioral Model}   \label{sec:modelingA}
We model the human motor learning behavior as the IOHMM~\cite{yin2017iohmm} illustrated in Fig.~\ref{fig:iohmm}.
The hidden state $h_k \in \{1, \dots, N\}$ represents one of $N$ discrete, latent motor skill states at trial $k$.
To maintain a reasonable number of parameters, the input vector $\mathbf{x}_k$ is composed of two experimental variables: the task type, captured by the slope angle $u_k$ of the target pair, and the haptic guidance, represented by the finger index $a_k$ to which the haptic feedback is applied (an index from $0\text{--}5$, where $0$ denotes no nudge, \rev{and $1-5$ denote fingers from thumb to the pinky}). A target pair consists of targets for trial $k$ and $k-1$, with their numbering convention following $\{0:(\mc T_2, \mc T_3),\ 1:(\mc T_1, \mc T_2),\ 2:(\mc T_1, \mc T_3),\ 3:(\mc T_1, \mc T_4),\ 4:(\mc T_2, \mc T_4),\ 5:(\mc T_3, \mc T_4)\}$.

The evolution of the latent motor skill state from trial $k$ to $k+1$ is governed by the transition probability distribution $\mathbb{P}(h_{k+1} | h_k, \mathbf{x}_k)$. The outputs $\mathbf{o}_k = [\texttt{RE}_k, \texttt{SoT}_k]\tran$ represent the performance metrics of the learner: Reaching Error (\texttt{RE}) and Straightness of Trajectory (\texttt{SoT}). The distribution of these outputs is conditional on the current state and the inputs, defined by the conditional distribution $\mathbb{P}(\mathbf{o}_k | h_k, \mathbf{x}_k)$. Together with the initial state distribution $\mathbb{P}(h_0)$, these components constitute the IOHMM of motor learning behavior under haptic guidance.

\subsection{Estimating the Model Parameters}
The parameters of the IOHMM were estimated using the experimental data collected from the \rev{first five} participants in the Heuristic Group \rev{performing the target capture task with heuristic nudging policy (see Section~\ref{sec:heuristic_policy},} and~\ref{sec:experiment_details}). For data-efficiency, functional approximation is used to estimate the state-transition and output probabilities.

\subsubsection{Initial and Transition Models}
The initial state distribution $\mathbb{P}(h_0 | \mathbf{x}_0)$ and the state transition probabilities $\mathbb{P}(h_{k+1} | h_k, \mathbf{x}_k)$ were modeled using Multinomial Logistic Regression (MNL)~\cite{kwak2002multinomial}. Formally, the probability of transitioning from a state $h_k = i$ to a state $h_{k+1} = j$ given input $\mathbf{x}_k$ is defined by the softmax function
\begin{equation}
    \mathbb{P}(h_{k+1}=j \mid h_k=i, \mathbf{x}_k) = \frac{\exp(\mathbf{w}_{ij}^\top \mathbf{x}_k + b_{ij})}{\sum_{l=1}^N \exp(\mathbf{w}_{il}^\top \mathbf{x}_k + b_{il})},
\end{equation}
where $\mathbf{w}_{ij}$ and $b_{ij}$ are the coefficient vector and intercept term associated with the transition $i \to j$. To prevent overfitting given the limited training data, we applied $L_2$ regularization (Ridge regression)~\cite{krogh1991simple} during the fitting process. The initial state distribution follows a randomized initialization since we do not enforce a semantic ordering of the latent states, and only perform a post-hoc re-ordering for visualization and interpretability.

\subsubsection{Output Model}
The outputs $\mathbf{o}_k$ (comprising \texttt{RE} and \texttt{SoT}) were modeled using a linear Gaussian function. For a given state $h_k = i$ and input $\mathbf{x}_k$, the observation is assumed to be drawn from a multivariate normal distribution where the mean is a linear function of the inputs
\begin{equation}
    \mathbb{P}(\mathbf{o}_k \mid h_k=i, \mathbf{x}_k) = \mathcal{N}(\mathbf{o}_k \mid \mathbf{V}_i \mathbf{x}_k + \mathbf{c}_i, \mathbf{\Sigma}_i),
\end{equation}
where $\mathbf{V}_i$ is the coefficient matrix, $\mathbf{c}_i$ is the intercept vector, and $\mathbf{\Sigma}_i$ is the covariance matrix for state $i$. To estimate the model parameters, the associated loss in the overall loss function is chosen as Ordinary Least Squares (OLS). To enhance model generalization, we applied Elastic-Net regularization~\cite{zou2005regularization}, which linearly combines $L_1$ and $L_2$ penalties, promoting sparsity while maintaining stability.

\subsubsection{Training Algorithm}
Since the dependencies in an IOHMM are parametric, the standard Baum-Welch algorithm cannot be applied directly. Instead, we utilized the Generalized Expectation-Maximization algorithm implemented in~\cite{bengio1996input}.

To mitigate the risk of convergence to local optima, we employed a Monte-Carlo (MC) training strategy. We executed multiple training runs, each utilizing one of $5$ random permutations of the subject data sequences. From these candidates, the model exhibiting the highest log-likelihood was selected. Empirical analysis on validation datasets confirmed that this permutation-based approach yielded models with superior generalizability compared to a single training sequence.

\subsubsection{Model Selection}
The model architecture and regularization weights were selected through K-fold cross-validation, maximizing the log-likelihood on validation data. This process yielded an optimal model size of $N=7$ latent states. The selected hyperparameters were $\alpha_{L_2}=1.0$ for the transition model, and $\alpha_{en}=10^{-4}$ with an $L_1$-ratio of $0.725$ for the emission model.
\begin{figure*}[t!]
    \centering
    \begin{subfigure}{0.5\linewidth}
        \centering
        \includegraphics[width=1\textwidth, height=1\textwidth, keepaspectratio]{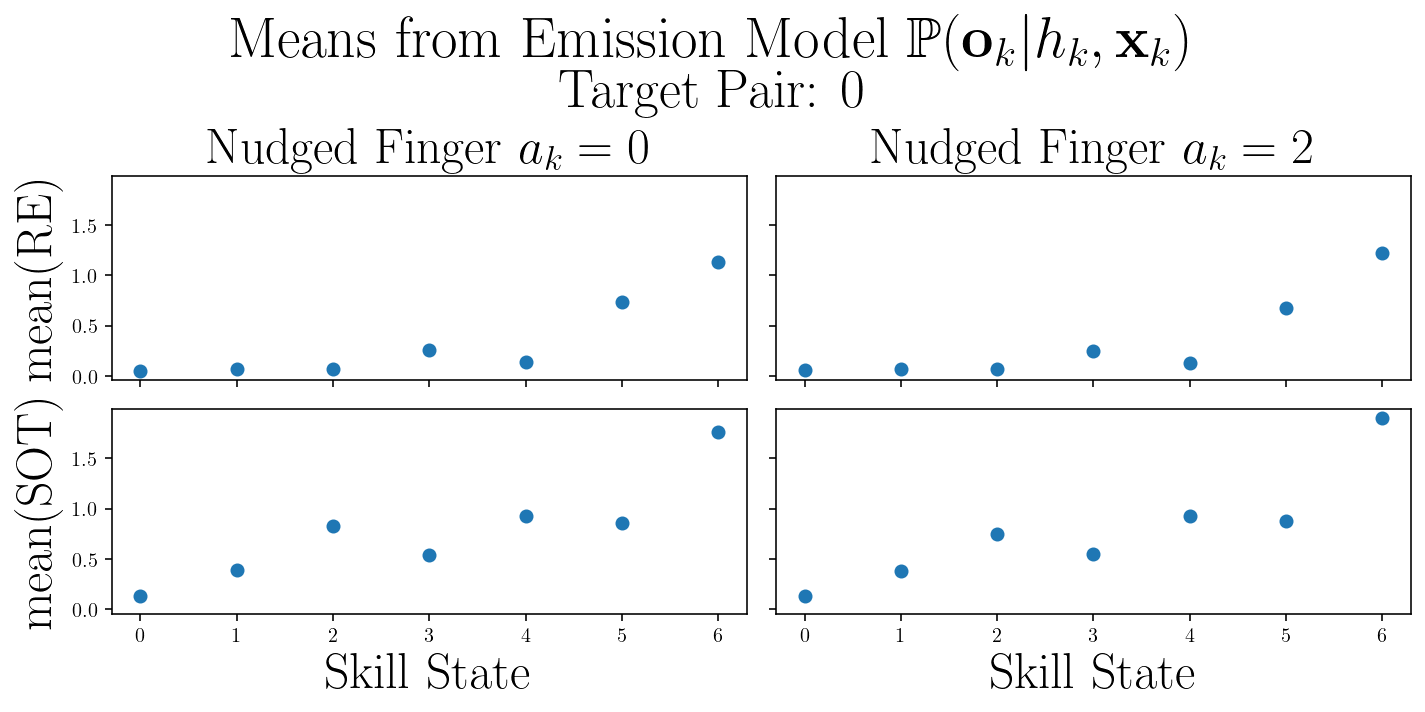}
        \subcaption{}
    \end{subfigure}%
    \begin{subfigure}{0.5\linewidth}
        \centering
        \includegraphics[width=1\textwidth, height=1\textwidth, keepaspectratio]{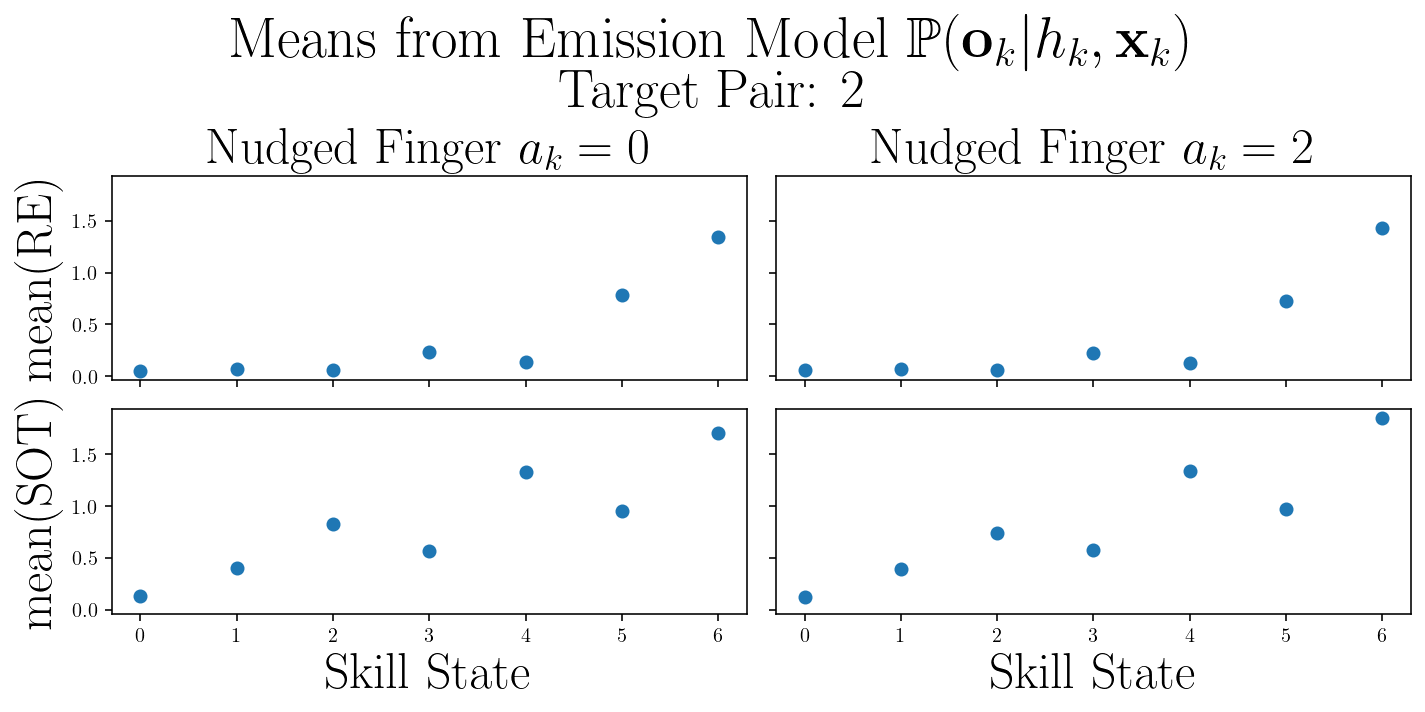}
        \subcaption{}
    \end{subfigure}%
    \\
    \centering
    \begin{subfigure}{0.5\linewidth}
        \centering
        \includegraphics[width=1\textwidth, height=1\textwidth, keepaspectratio]{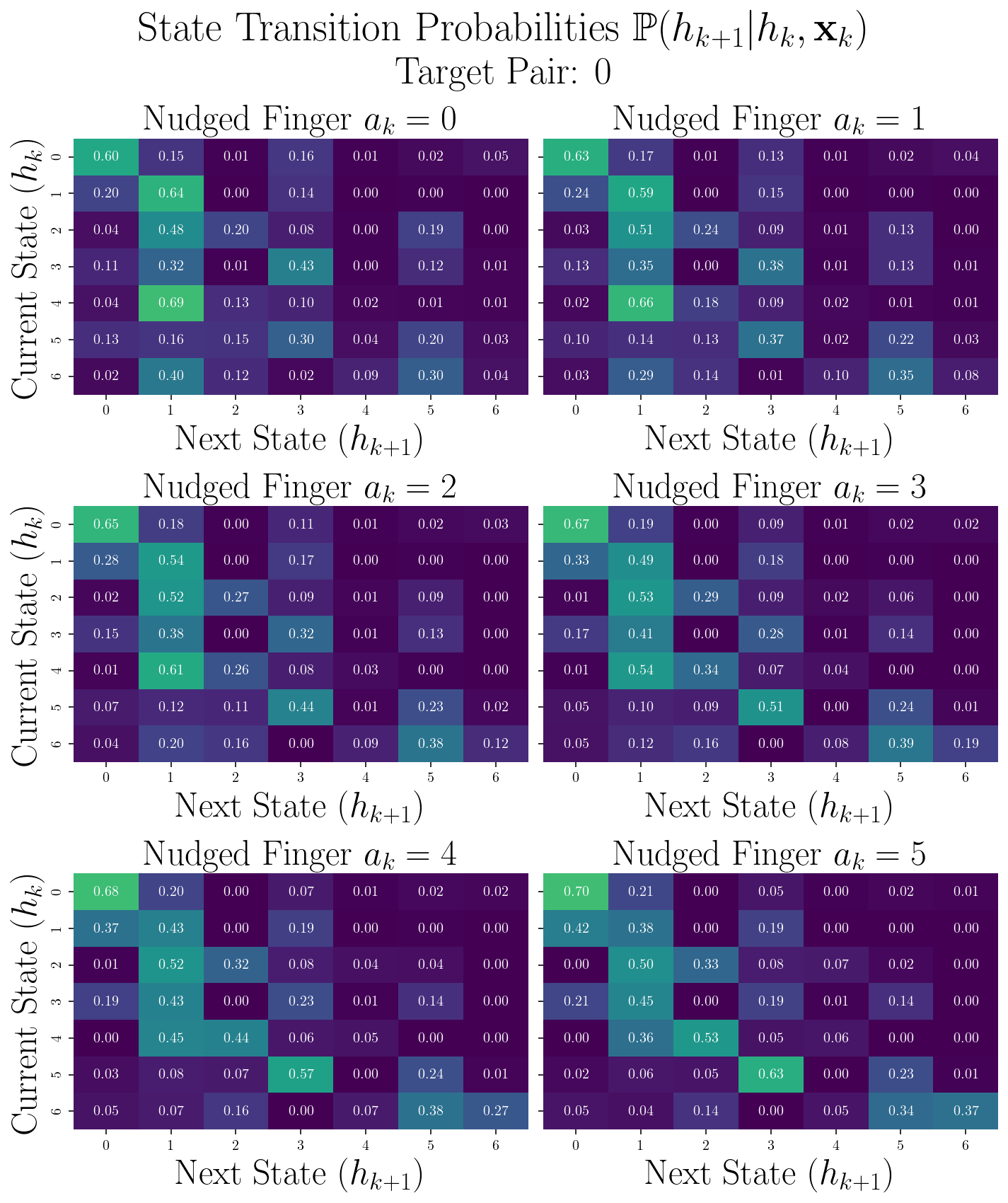}
        \subcaption{}
    \end{subfigure} \hspace{-3em}%
    \begin{subfigure}{0.5\linewidth}
        \centering
        \includegraphics[width=1\textwidth, height=1\textwidth, keepaspectratio]{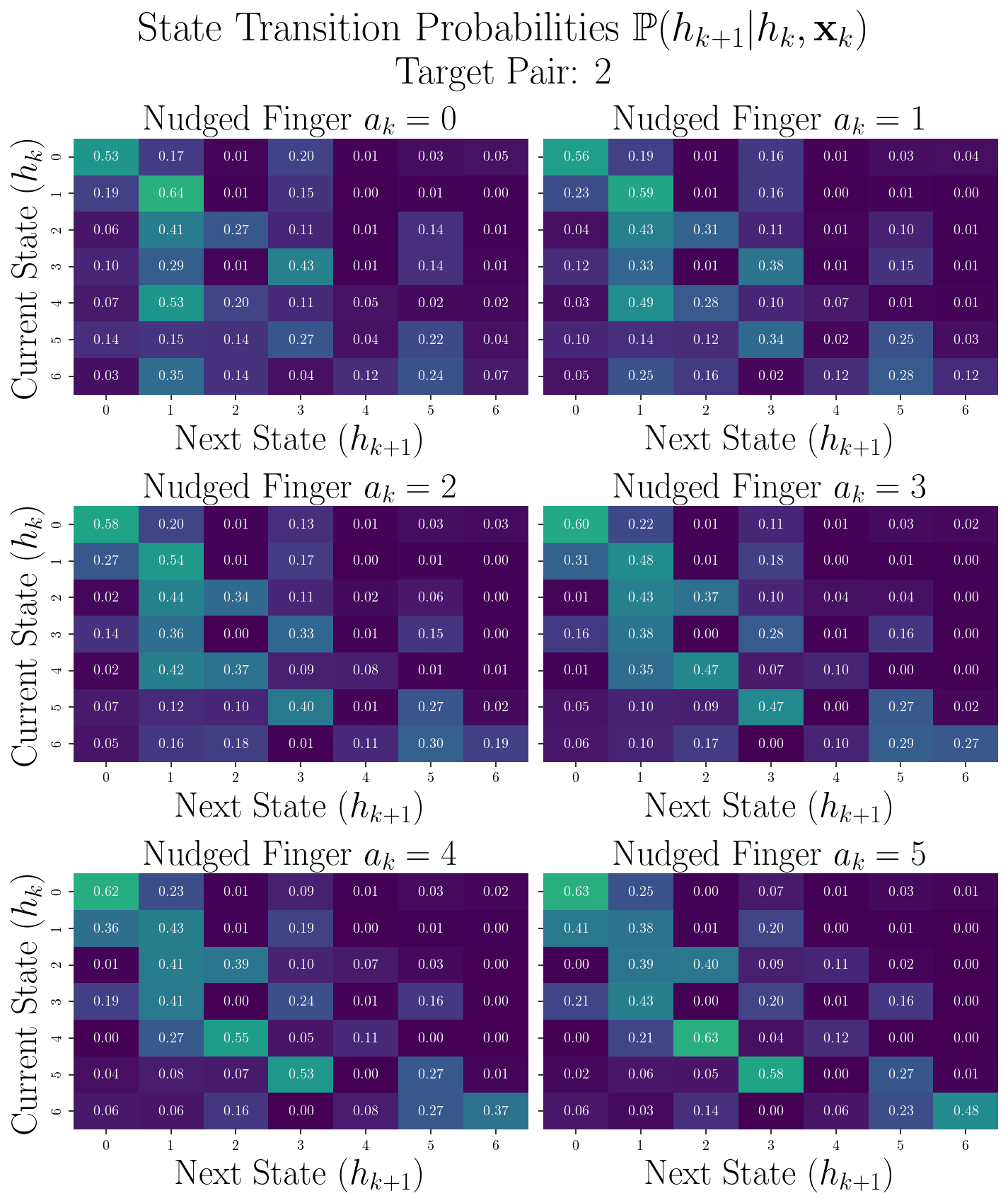}
        \subcaption{}
    \end{subfigure}
    \caption{\textbf{Estimated Behavioral Model:} (a) Predicted \texttt{RE} and \texttt{SoT} means out of the emission model for the ordered skill states, and (b) State Transition Matrices extracted from the estimated human motor learning behavioral model show how participants tend to transition from low latent skill states to high latent skill states under the effect of the input slope angles $u_k = -1.57, -1.11$  (corresponding to target pairs $\{0, 2\}$) and the nudged finger indices $a_k = {0,\cdots,5}$. The latent skill states are ordered best-to-worst $(0-6)$ based on their mean output $\texttt{RE}$ emission values. The $\texttt{nudge} = 0$ value represents no nudging.}
    \label{fig:STM}
\end{figure*}
\subsection{State Transition Matrices} \label{sec:STM}
\rev{We now analyze the motor learning behavior of the participants subject to task type and the nudge feedback captured by the estimated IOHMM model.}
Accordingly, to study how their latent skill states probabilistically transition as a function of target pair and nudge feedback, we plot the transition function as state transition matrices for the pair of inputs $\mathbf{x}_k$. 
For interpretability, we first order the latent motor skill states from best $(0)$ to worst $(6)$ based on the increasing value of the average of predicted $(\text{mean}(\texttt{RE})+\text{mean}(\texttt{SoT}))$ outputs for all inputs (Fig.~\ref{fig:STM}).
We then analyze these transitions for the hardest and easiest target pair, denoted by their slopes, under the effect of different nudge feedbacks, where $0$ refers to no nudge. The order of difficulty of various target pairs is obtained through studying the number of experiment trials it takes for \texttt{SoT} to converge to a baseline value of $0.3$ on average across the five participants whose data is used to estimate the IOHMM. Target pair $0: (\mc T_2, \mc T_3)$ with slope angle $\{\pm 1.57\}$ is the easiest with participants converging to an \texttt{SoT} of $0.3$ by trial $19$ on average, whereas target pair $2: (\mc T_1, \mc T_3)$ with slope angle $\{-1.11, 2.03\}$ is the hardest, with \texttt{SoT} converging to $0.3$ by trial $210$ on average.
Fig.~\ref{fig:STM} shows the state transition matrices corresponding to the target pair $0$ and $2$.

The probabilities show how participants tend to transition from low-performing motor skill states to high-performing motor skill states, imparting a lower triangular dominant structure to these matrices owing to the mean performance-based skill ordering.

The figure also reveals that for both the target pairs, the participants tend to stay in high performing skill states $\{0,1\}$ without needing any nudge. Additionally, for the target pair $0$, a nudge to finger $1$, which is the thumb, helps them transition from lower to higher performing skill states more prominently. This was also observed in the experiments, where the thumb abduction-adduction majorly led to the screen cursor's vertical motion. 
Furthermore, Fig.~\ref{fig:STM} also shows that for the target pair $2$, nudges are more helpful for transitioning from lower to higher performing skill states compared to nudges for target pair $0$. \rev{It should be noted that the observed nudging effect with respect to different finger indices are dependent on the forward BoMI mapping matrix $C$ constructed via the consistent calibration performed for all participants.}

\section{Skill-based Nudging Prescription Design}     \label{sec:pomdp_design}
Building on the premise that true motor skill is latent and high-dimensional tasks lack unique reference trajectories, we shift from conventional performance-error correction to skill-state-aware guidance. In this section, we first outline a heuristic nudging policy aimed at local kinematic refinement. Subsequently, leveraging the probabilistic behavioral model of motor learning (detailed in Section~\ref{sec:IOHMM}), we formulate a POMDP to synthesize optimal, personalized haptic nudges designed to regulate motor exploration, resolve kinematic ambiguity, and accelerate the discovery of efficient coordination strategies.

\subsection{Heuristic Nudging Policy}   \label{sec:heuristic_policy}
We synthesized a heuristic feedback policy informed by pilot studies. \rev{This strategy aims to assist the user by incrementally refining their hand configuration to achieve the current objective by nudging the finger capable of the largest performance error reduction with the least change in finger pose.} For a specific target $t_k$ presented in trial $k$, we first calculate a theoretical optimal posture, $\supscr{\theta}{opt}_k$, by applying the pseudo-inverse of the BoMI mapping matrix, such that $\supscr{\theta}{opt}_k = C^{\dagger}t_k$.
We then evaluate the difference between this ideal configuration and the participant's actual hand posture, $\supscr{\theta}{curr}_k$, at the onset of the trial. The deviation of each finger $i$ is quantified in joint space as the Sum of Squared Errors (SSE) between the current and optimal angles as
\begin{equation}
    d_{i,k} = \sum_{j=1}^{4} \left(\supscr{\theta_{ij,k}}{curr} - \supscr{\theta_{ij,k}}{opt}\right)^2,
\end{equation}
where $\supscr{\theta_{ij,k}}{curr}$ represents the angle of the $\supscr{j}{th}$ joint of the $\supscr{i}{th}$ finger at the start of trial $k$. To prioritize fingers that require minimal adjustment to correct the cursor position, we invert these distance metrics to generate a closeness score. These scores are then normalized into a probability distribution via a \texttt{softmax} function with a temperature parameter $\tau$ as
\begin{equation}
    \mb P(a_k=i) = \frac{e^{d_{i,k}^{-1}/\tau}}{\sum_{n=1}^5 e^{d_{n,k}^{-1}/\tau}}.
\end{equation}
The specific finger to receive the haptic nudge is sampled from this distribution. Here, minimizing $d_i$ identifies the finger capable of producing the most significant cursor position with the least kinematic effort. Consequently, this heuristic functions as a local refinement mechanism. It serves as a computationally lightweight standard against which we evaluate our optimal, POMDP-driven nudging policy.

\subsection{POMDP-based Optimal Nudging Policy}
To synthesize an active intervention strategy, we treat the nudge input in the IOHMM as a control variable and design a control policy using a POMDP framework.
This policy $\pi$ maps the posterior distribution over human motor skill state (called belief) to specific haptic nudge feedback.
\\
\textbf{State Space, Action Space, and Observations:}
The POMDP state is defined as the tuple $s_k = (h_k, t_{k-1}, t_k) \in \mathcal{S}$, which encapsulates the latent skill level $h_k$ alongside the previous target $t_{k-1}$ and the current active target $t_k \in \mathcal{T} \backslash\{t_{k-1}\}$. The action variable $a_k = \texttt{nudge}_k \in \mathcal{A}$ represents the index of the nudged finger, and the observation space $o_k \in \Omega$ remains consistent with the IOHMM formulation.
\\
\textbf{Transition and Output Models:}
Given that at target $t_{k-1}$, the next target $t_{k}$ is drawn uniformly randomly from $\mc T\backslash t_{k-1}$ and is independent of the motor state, the composite transition model is written as
\begin{equation}
    \mb P(s_{k+1} | s_k, a_k) = \mb P(h_{k+1}|h_k, u_k, a_k) \times \mb P(t_k|t_{k-1}).
\end{equation}
Similarly, the output model $\mb P(o_k| h_k, u_k, a_k )$ is inherited directly from the IOHMM structure
It is important to note that $u_k$ is completely determined from the target pair $\{t_{k-1}, t_k\}$.
\\
\textbf{Reward Function:}
To foster both immediate task success and long-term motor learning, the designed reward function $R(s_k, a_k)$ balances instantaneous performance against future skill acquisition potential. Mathematically, this is defined as the negative sum of the expected immediate cost and the generalized utility of the next latent state as
\begin{align}
    R(s_k, a_k) = -&\left(\subscr{C}{imm}(h_k, u_k, a_k) \right. \notag \\
    &\left. + \subscr{w}{g} \subscr{C}{gen}(h_{k+1}, h_k, u_k, a_k)\right),
\end{align}
where $\subscr{C}{imm}(s_k, a_k)$ denotes the cost for the current trial, while $\subscr{C}{gen}(h_{k+1}, s_k, a_k)$ evaluates the expected performance of all possible inputs over all possible next latent states $h_{k+1}$. The weighting parameter was fixed at $\subscr{w}{g} = 1$. The immediate performance penalty is a linear combination of the predicted means for \texttt{RE} and \texttt{SoT} as
\begin{align}
    \subscr{C}{imm} = \subscr{\mu}{\texttt{RE}}(h_k,u_k,a_k) + \subscr{w}{\texttt{SoT}} \subscr{\mu}{\texttt{SoT}}(h_k,u_k,a_k),
\end{align}
where $\subscr{\mu}{\texttt{RE}}$ and $\subscr{\mu}{\texttt{SoT}}$ are the predicted \texttt{RE} and \texttt{SoT} means from the output model, and $\subscr{w}{\texttt{SoT}} = 2$ scales the relative importance of \texttt{SoT}.
\rev{The second term in the reward $\subscr{C}{gen}$ captures the versatility of the skill state at stage $k+1$, and is defined by
\begin{equation}
    \subscr{C}{gen} = \expt_{h_{k+1}} [\mb P(h_{k+1}|h_k, u_k, a_k) \bs\cdot \subscr{C}{g}(h_{k+1})],
\end{equation}
where $(\bs \cdot)$ denotes the dot product. The term $\subscr{C}{g}$ serves as a heuristic for skill versatility. It quantifies the inherent value of latent skill state $h_{k+1}$
as
\begin{equation}
    \subscr{C}{g}(h_{k+1}) = \sum_{t_{k}} \sum_{t_k+1{} \neq t_{k}} \sum_{a_{k+1}} \subscr{C}{imm}(h_{k+1}, u_{k+1}, a_{k+1}).
\end{equation}
}
By minimizing this composite cost, the planner is encouraged to select haptic nudge indices that not only reduce current errors but, critically, steer the learner toward skill states with strong performance across all possible tasks.

\subsection{Approximate POMDP Solution via QMDP}
\rev{A POMDP can be interpreted as an MDP in the belief space, a continuous, high-dimensional simplex representing the set of all probability distributions over the underlying POMDP states $s_k \in \mc S$.
The belief state $b_k:=b(s_k)$ represents the posterior probability distribution $\mb P(s_k | o_k, a_k, s_{k-1})$ and is iteratively computed given observation $o_k$ as
\begin{equation} \label{eq:belief_update}
    b_{k+1} =  \eta \mb P(o_k | s_{k}, a_k) \sum_{s \in \mc S} \mb P(s_{k+1}| s_k, a_k) b_k,
\end{equation}
where $\eta$ is the normalization constant.
Given the computational intractability of applying standard infinite-horizon value iteration directly to this continuous belief space, we adopt the QMDP approximation method~\cite{littman1995learning} to obtain a near-optimal nudging policy. QMDP simplifies the problem by solving the underlying MDP, under the assumption of full state observability, to obtain the optimal action-value function $\subscr{Q}{MDP}(s_k, a_k)$ by iteratively solving
\begin{align}   \label{eq:qmdp}
    \subscr{Q}{MDP}(s_k,a_k) &= R(s_k,a_k) + \gamma \sum_{s_{k+1} \in \mathcal{S}} \mb P(s_{k+1}|s_k,a_k)V(s_{k+1}), \notag \\
    V(s_k) &= \max_{a\in \mc A} \subscr{Q}{MDP}(s_k, a),
\end{align}
until the value function $V(s_k)$ converges across the state space.
The QMDP algorithm then samples a state from the posterior belief and applies the associated optimal action.
To further robustify the policy against model imperfections stemming from limited training data, the MDP is solved via infinite-horizon soft-value iteration~\cite{haarnoja2018soft}. The value update rule follows the soft Bellman update wherein the maximum operator in Bellman equation~\eqref{eq:qmdp} is replaced with a softmax operator with parameter $\alpha$ as
\begin{equation}
    V(s_k) \leftarrow \alpha \log \sum_{a \in \mc A} \exp\left(\frac{\subscr{Q}{MDP}(s_k,a)}{\alpha}\right).
\end{equation}
Small values of $\alpha$ correspond to near optimality, while larger values correspond to a random action selection. This parameter acts as a regularizer and guards against model imperfections.
The final output is a stochastic policy $\pi^*(a|s_k)$, where action $a$ is selected with probability
\begin{equation}
    p(\subscr{Q}{MDP}(s_k,a)) \propto \frac{\exp\left(\subscr{Q}{MDP}(s_k,a)/\alpha\right)}{\sum_{a'}\exp\left(\subscr{Q}{MDP}(s_k,a')/\alpha\right)}.
\end{equation}
For this implementation, we used $\alpha=0.2$ and $\gamma=0.98$. These values were obtained through Monte Carlo simulations of the experiment, using the trained IOHMM as a generative model to maximize the mean policy performance.}

\subsubsection*{Real-Time Policy Execution}
The derived stationary policy is integrated into the experiment for real-time execution.  Throughout the experiment, the system maintains a running belief state $b_k$.
The selection of the specific finger index for haptic feedback proceeds in two stages: first, a state hypothesis $s_k$ is sampled from the current belief distribution $s_k \sim b_k$; second, the action $a_k$ is sampled from the policy distribution $\pi^*(a|s_k)$ corresponding to the sampled state. Following the prescription of the haptic nudge by the BoMI, the system records the participant's trial performance metrics $o_k = \{\texttt{RE}_k, \texttt{SoT}_k\}$. The belief state is then advanced to the subsequent time step via the belief update~\eqref{eq:belief_update}. This updated belief $b_{k+1}$ encapsulates the new system knowledge and is used in the decision process for the next trial.
\begin{figure}[t!]
    \centering
    \begin{subfigure}{0.85\linewidth}
	    \centering
        \caption{}
        \includegraphics[width=1\linewidth, height=1\linewidth, keepaspectratio]{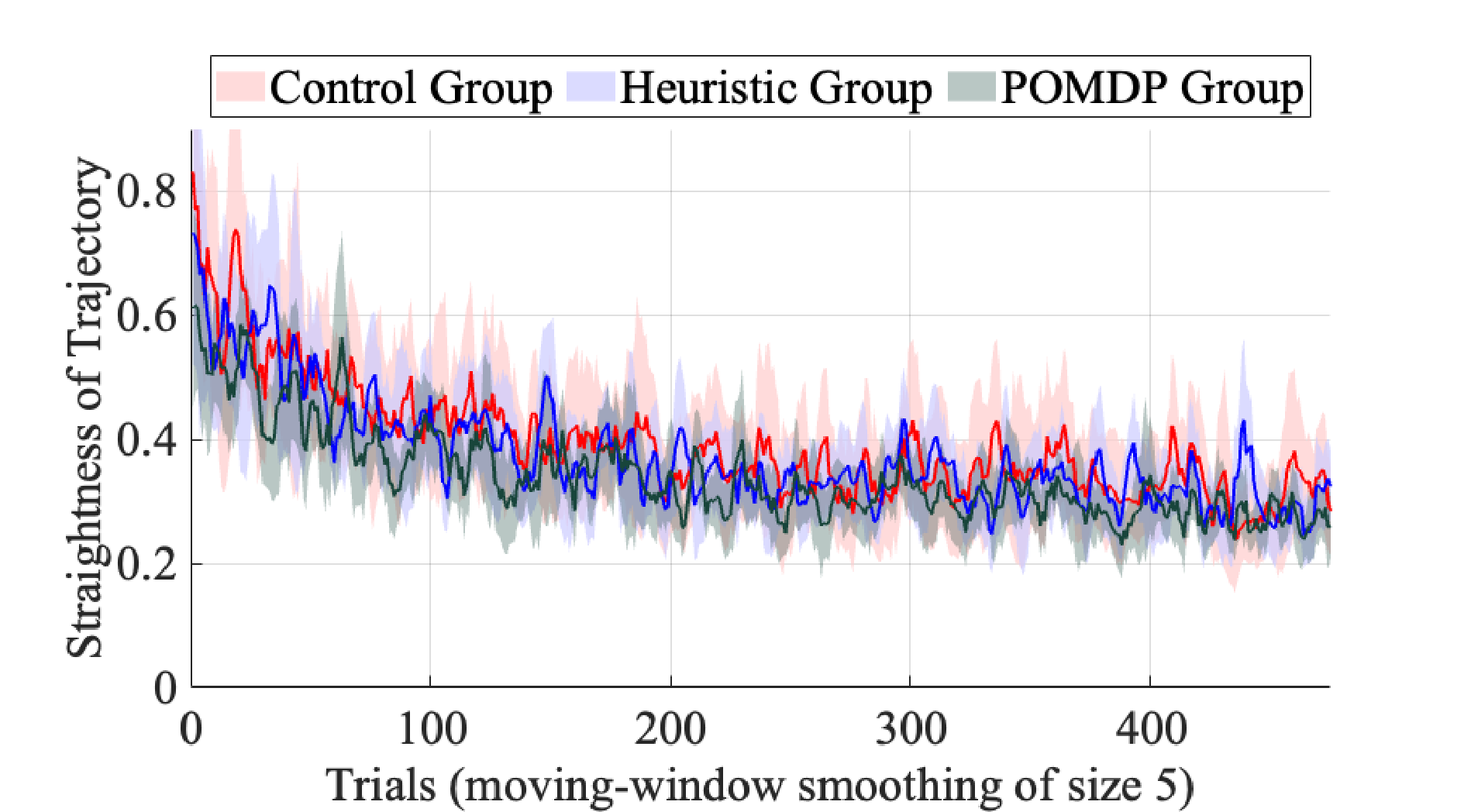}
        \label{fig:SOT_curves}
    \end{subfigure}\vspace{-1.2em}%
    \\
    \centering
    \begin{subfigure}{0.85\linewidth}
	    \centering
        \caption{}
        \includegraphics[width=1\linewidth, height=1\linewidth, keepaspectratio]{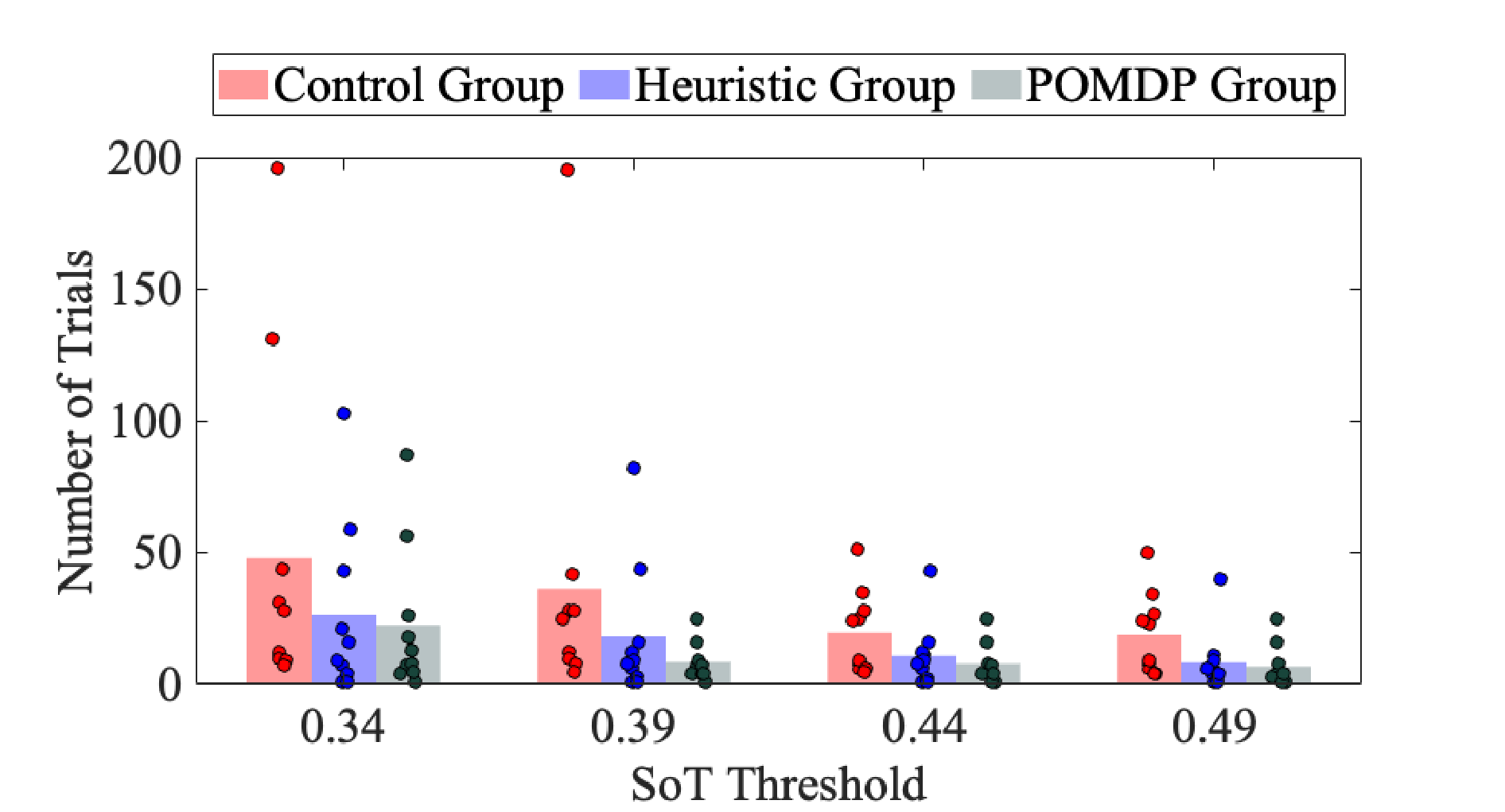}
        \label{fig:SOT_threshold}
    \end{subfigure}\vspace{-1.2em}%
    \caption{\textbf{Movement Efficiency:} (a) Mean \texttt{SoT} curves (with $95\%$ confidence intervals) of the three group participants, and (b) the mean number of trials (along with the scatter plot taken across participants) to converge to a specific \texttt{SoT} threshold under the three nudging policies shows fastest convergence for participants trained on POMDP-based optimal nudging policy, followed by the heuristic policy. 
    The low p-values from two-tailed tests from the linear mixed model fits show a decrease in \texttt{SoT} across trials and in the number of trials required by the POMDP group for various \texttt{SoT} thresholds. Each scatter plot point represents the number of trials taken by a participant to achieve the \texttt{SoT} threshold.
    }
    \label{fig:exp_results}
    \vspace{-0.2in}
\end{figure}
\section{Results}   \label{sec:results}
This section presents the findings from our human-subject study, designed to evaluate the efficacy of our proposed optimal nudging policy
compared to a no-feedback baseline and a standard heuristics-based strategy.

\subsection{Experiment Details}     \label{sec:experiment_details}
To validate our approach, we divided participants into three distinct experimental groups, all of whom performed the target capture task described in Section~\ref{sec:target_capture_game}. These groups were defined as follows: a Control Group, which received no haptic feedback; a Heuristic Group, which received feedback based on the heuristic nudge feedback policy; and a POMDP Group, which received our optimal, skill-state-aware nudge feedback. Each group consisted of $10$ participants\footnote{All human subject experiments were approved under the Michigan State University Institutional Review Board Study ID LEGACY14-431M.}.
The experimental timeline was structured sequentially. First, data was collected from the Heuristic Group. The input-output experiment data from the initial five participants in this group were utilized to train the Input-Output Hidden Markov Model (IOHMM), establishing the baseline behavioral model of human motor learning.
For the POMDP Group, the haptic feedback was generated by casting the problem as a POMDP and solving it approximately via the QMDP method. During the experiment, the system maintained a running belief state, which was updated after every trial. To select an optimal action for the subsequent trial, a state was sampled from this current belief distribution and mapped to a specific nudge finger index via the QMDP policy. The belief state was initialized using the prior distribution $\mb P(h_0)$ derived from the trained IOHMM.
\subsection{Movement Efficiency: Straightness of Trajectory}
We used the Straightness of Trajectory \texttt{SoT} to measure movement efficiency across the three groups, where a lower value signifies a smoother, more desirable movement.
We conducted two primary analyses to compare the three participant groups. First, we tracked the evolution of \texttt{SoT} across all experiment trials to assess the overall performance trajectories. Second, we computed the number of trials required for participants to converge to a predefined \texttt{SoT} threshold. This convergence analysis directly evaluates the efficiency of a nudging policy, as haptic feedback can nudge participants towards efficient joint movements that could improve task performance and thus require fewer trials to reach a performance threshold.
Moreover, a nudging policy based on the motor skill state of the participant could help achieve accelerated motor skill acquisition by optimally nudging the right joints towards better synergistic patterns, going beyond simple task performance improvements.
\rev{To analyze performance across the full learning trajectory, we fit a Linear Mixed Model (LMM) to the \texttt{SoT} values for all $480$ experiment trials from each participant, with trial number, group, and their interaction as fixed effects, and a random intercept per participant, estimated via Restricted Maximum Likelihood (REML)~\cite{corbeil1976restricted}. This model accounts for the correlation structure arising from repeated measurements within participants. Post-hoc two-tailed contrasts with Bonferroni correction were used to test for pairwise differences between the POMDP, heuristic, and control groups. For the threshold-based analysis, where each participant contributes a single scalar value (the trial index at which a given \texttt{SoT} threshold is first reached), we conducted a one-way ANOVA to test for an overall group effect, followed by pairwise independent-samples t-tests (two-tailed) with Bonferroni correction.}

The results from Fig.~\ref{fig:exp_results} show that \texttt{SoT} decreases with trials for all groups, and the mean number of trials taken to achieve a particular \texttt{SoT} threshold also decreases for increasing thresholds, indicating task learning improvement through training. The p-values and Cohen's d-values from Table~\ref{table:SOT_pvals} show that participants trained on the optimal nudging policy achieved lower \texttt{SoT} values on average across the experiment trials as compared to participants trained with no haptic nudges and heuristic nudges; however, the results are not statistically significant. Furthermore, the POMDP group takes statistically significantly (at significance level $\alpha = 0.10$) fewer trials than the control group to converge to higher \texttt{SoT} thresholds. This indicates expedited performance improvement in the early stages of training.
Although the POMDP group converged to various \texttt{SoT} thresholds earlier than the heuristic group on average, the results are not statistically significant.
\begin{table}[b!]
    \centering
    \caption{\textbf{Two-tailed Contrasts for \texttt{SoT} Group Differences.} The p-values for the hypothesis that the POMDP group has different \texttt{SoT} values across gameplay trials and different number of trials to achieve thresholds compared to the control and the heuristic group are reported. The results show that the POMDP group achieves lower \texttt{SoT} values on average, and takes statistically significantly (at significance level $\alpha=0.10$) fewer gameplay trials to achieve \texttt{SoT} thresholds early in the training.
    }
    \begin{tabular}{||c|c||c|c|c|c||}
        \hline
        \multicolumn{2}{||c||}{\multirow{2}{*}{\textbf{Metrics}}}&
        \multicolumn{2}{c|}{\textbf{POMDP vs Control}} & \multicolumn{2}{c||}{\textbf{POMDP vs Heuristic}} \\
        \cline{3-6}
        \multicolumn{2}{||c||}{}& \textbf{$\subscr{p}{bonf}$} & \textbf{$\subscr{d}{cohen}$}& \textbf{$\subscr{p}{bonf}$} & \textbf{$\subscr{d}{cohen}$} \\
        \hline \hline
        \multicolumn{2}{||c||}{\textbf{SoT Curve}} & 0.1053 & -0.5368 & 0.5002 & -0.6048 \\
        \hline
         & 0.34 & {0.5290} & -0.5151 & 1.0000 & -0.1277\\
        \cline{2-6}
        \textbf{SoT} & 0.39 & {0.2865} & -0.6845 & 1.0000 & -0.5148\\
        \cline{2-6}
        \textbf{Thresholds} & 0.44 & \textbf{0.0875} & -0.9698 & 1.0000 & -0.2950\\
        \cline{2-6}
        & 0.49 & \textbf{0.0796} & -0.9912 & 1.0000 & -0.1619\\
        \hline
    \end{tabular}
    \label{table:SOT_pvals}
\end{table}
\begin{figure}[ht!]
    \vspace{-0.2in}
    \centering
    \begin{subfigure}{0.85\linewidth}
	    \centering
        \caption{}
        \includegraphics[width=1\linewidth, height=1\linewidth, keepaspectratio]{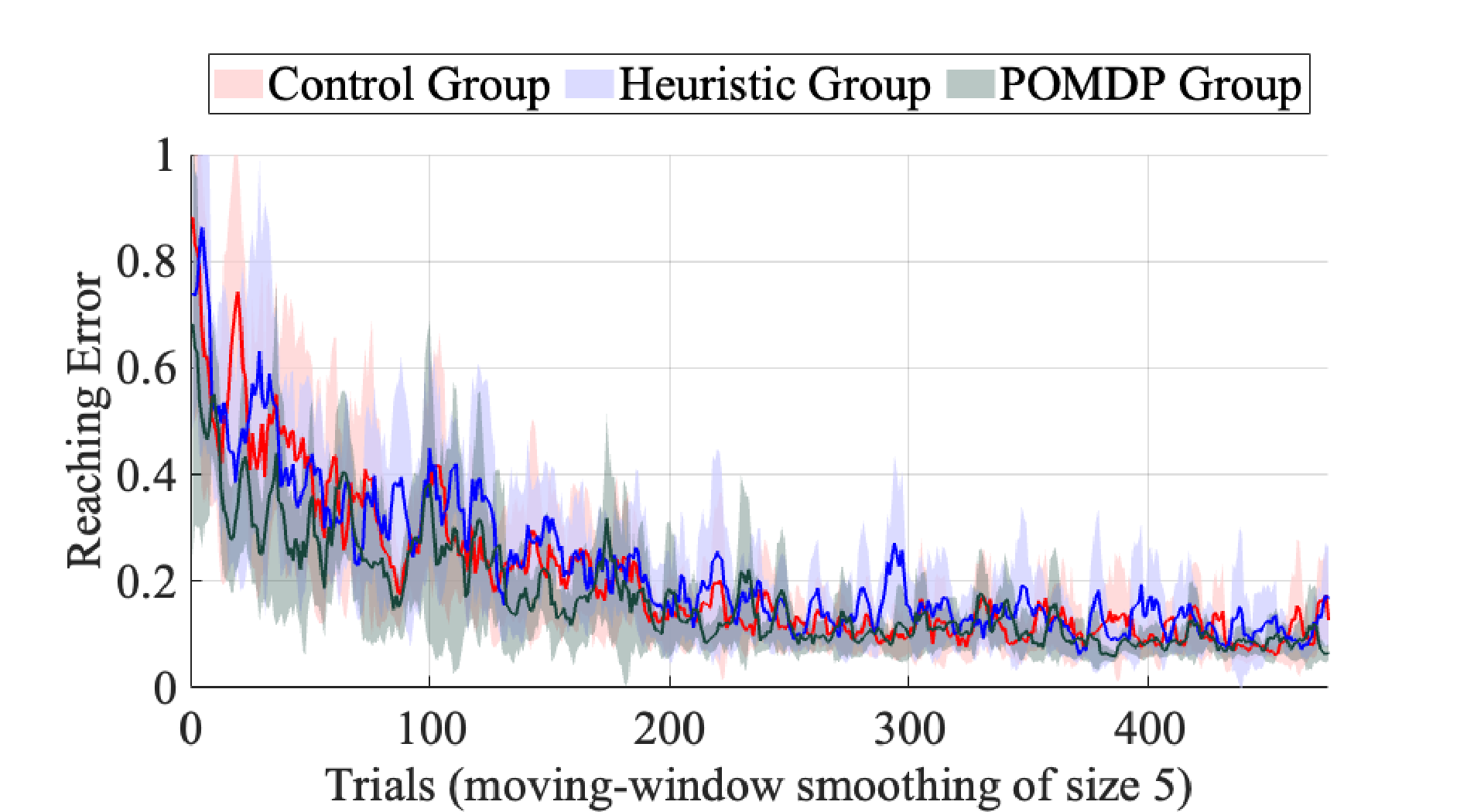}
        \label{fig:RE_curves}
    \end{subfigure}\vspace{-1.2em}%
    \\
    \centering
    \begin{subfigure}{0.85\linewidth}
	    \centering
        \caption{}
        \includegraphics[width=1\linewidth, height=1\linewidth, keepaspectratio]{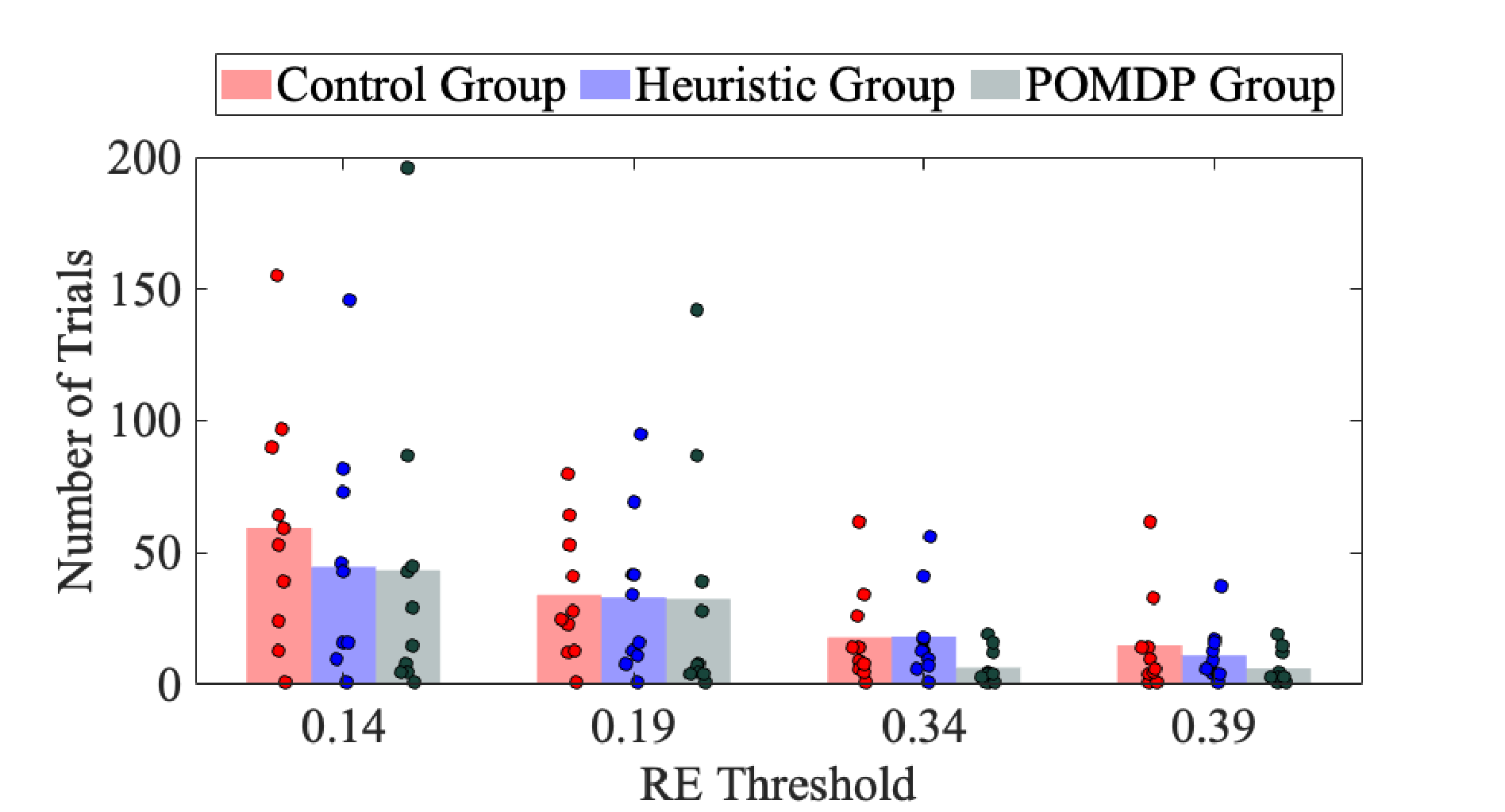}
        \label{fig:RE_threshold}
    \end{subfigure}\vspace{-1.2em}%
    \caption{\textbf{Endpoint Accuracy:} (a) Mean \texttt{RE} curves (with $95\%$ confidence intervals) of the three group participants, and (b) the mean number of trials (along with the scatter plot taken across participants) to converge to different \texttt{RE} thresholds under the three nudging policies shows fastest convergence for participants trained on POMDP-based optimal nudging policy. 
    The low two-tailed test p-values from the linear mixed model fits show a decrease in \texttt{RE} across trials and in the number of trials required by the POMDP group for all \texttt{RE} thresholds. Each scatter plot point represents the number of trials taken by a participant to achieve that \texttt{RE} threshold.
    }
    \label{fig:RE_results}
    \vspace{-0.2in}
\end{figure}
\begin{figure*}[h!]
    \centering
    \begin{subfigure}{0.5\linewidth}
	    \centering
        \caption{}
        \includegraphics[width=1\linewidth, height=1\linewidth, keepaspectratio]{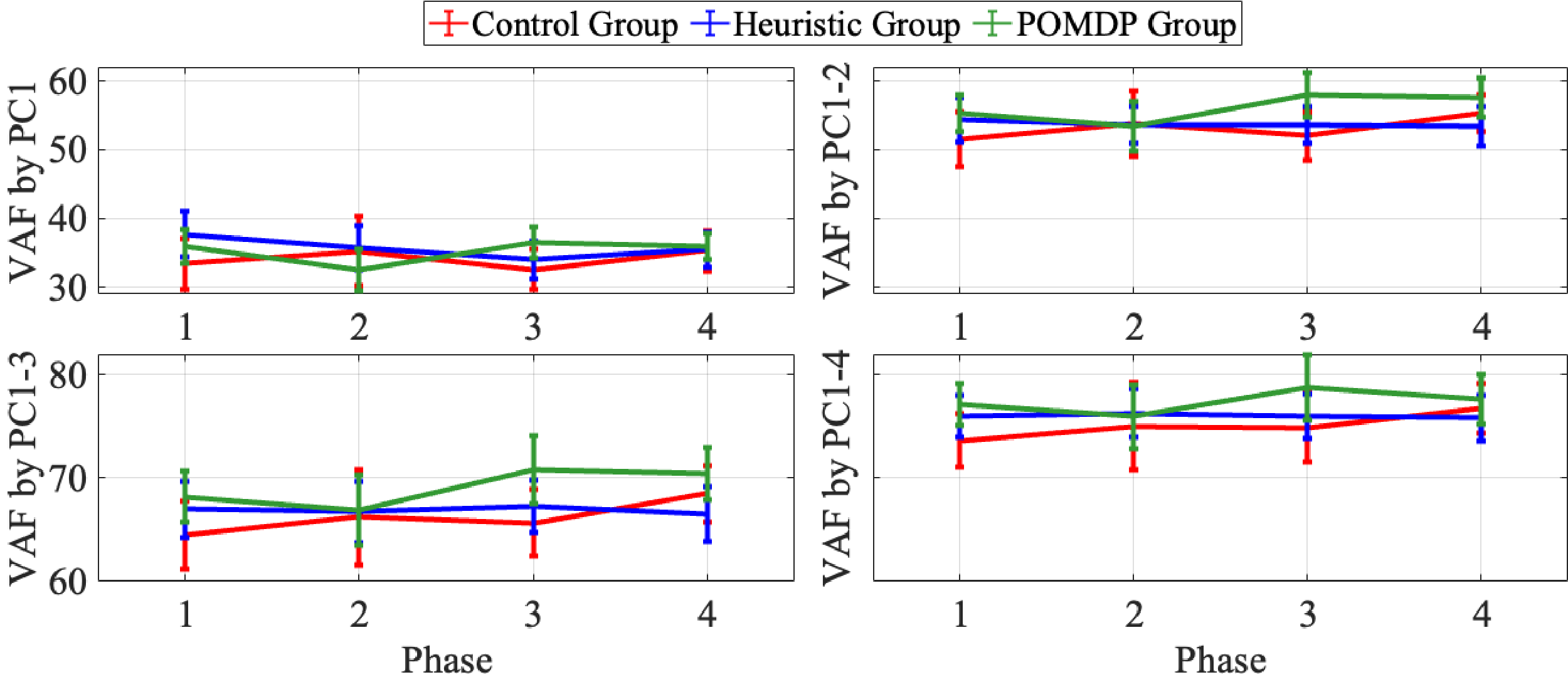}
        \label{fig:vaf}
    \end{subfigure}\hspace{-1.2em}%
    ~
    \centering
    \begin{subfigure}{0.4\linewidth}
	    \centering
        \caption{}
        \includegraphics[width=1\linewidth, height=1\linewidth, keepaspectratio]{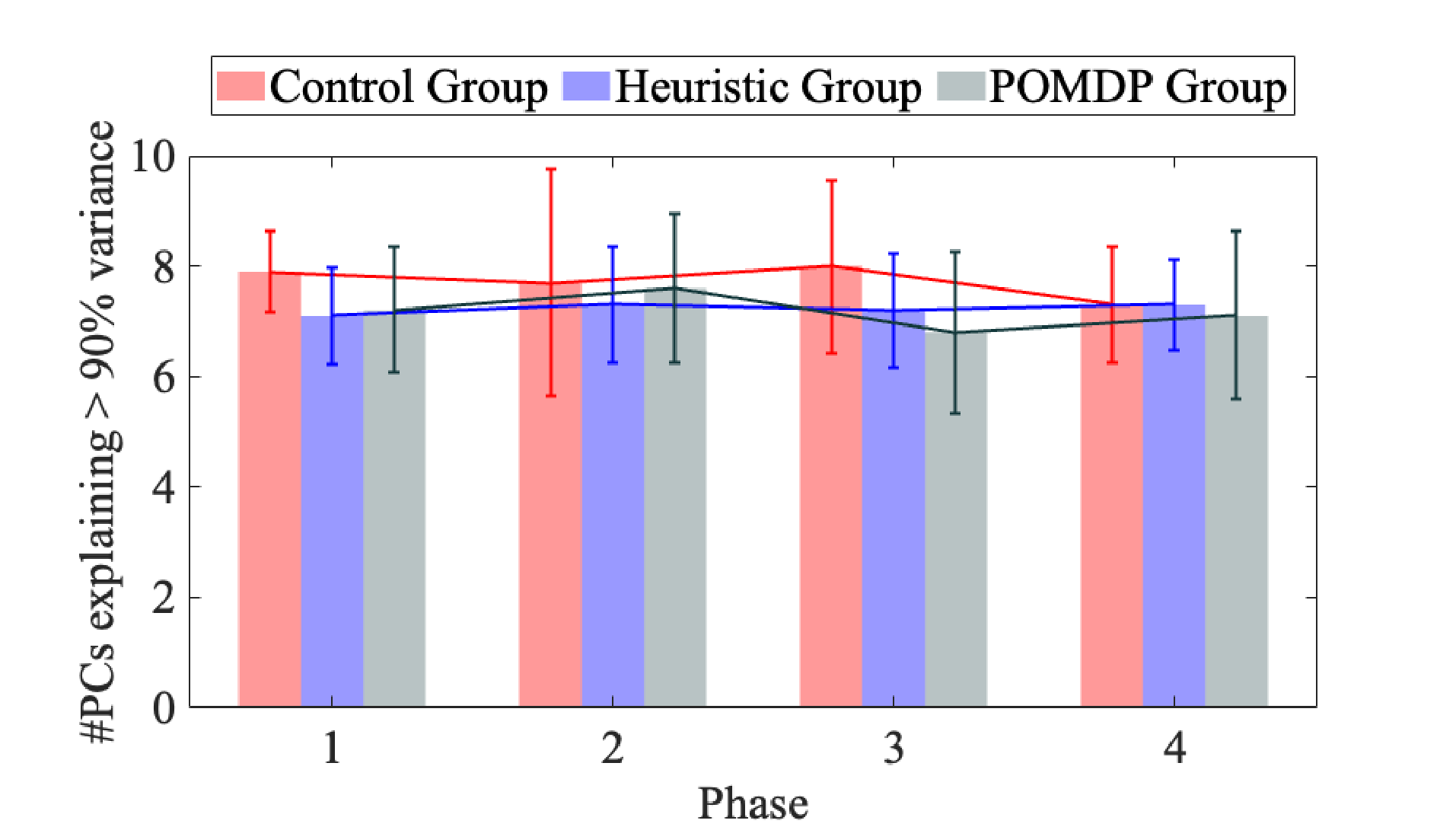}
        \label{fig:pc90}
    \end{subfigure}\vspace{-1.2em}%
    \caption{\textbf{Synergy Analysis:} (a) Variance accounted for (VAF) by the first four principal components (PC), and (b) the number of PCs required to explain more than $90\%$ variance in joint data across the three participant groups reveal that the POMDP group achieved a higher VAF by top-PCs (especially in later phases) and required fewer PCs to explain $>90\%$ variance in the data on average, signifying a more rapid discovery of the task relevant subspace towards expedited skill acquisition. Training blocks are consolidated into phases, with each phase consisting of two consecutive blocks.}
    \label{fig:synergy_analysis}
    \vspace{-0.1in}
\end{figure*}
\subsection{Endpoint Accuracy: Reaching Error}
We used the reaching error \texttt{RE} to measure the endpoint accuracy of the three experiment groups. Recall that the \texttt{RE} for an experiment trial is the Euclidean distance of the screen cursor from the trial target at the end of $2~$s, or end of movement, whichever is earlier. Thus, lower \texttt{RE} values signify better task performance. We performed two analyses: first, we studied the evolution of \texttt{RE} across experiment trials to assess overall motor performance, and second, we tracked the number of trials required for participants to converge to predefined \texttt{RE} thresholds.

Results from Fig. \ref{fig:RE_results} show that the \texttt{RE} decreases across all groups as the experiment progresses, indicating improved motor performance with training. The POMDP group achieves statistically significantly (at significance level $\alpha<0.10$) lower \texttt{RE} compared to the other two groups across all experiment trials, as shown in Table~\ref{table:RE_pvals}. Results also show that the POMDP group requires fewer trials on average to converge to lower \texttt{RE} thresholds (Fig.~\ref{fig:RE_threshold}) than the control and the heuristic group, indicating expedited motor performance improvement; however, the results are not statistically significant.
\subsection{Synergy Analysis}
The inherent high redundancy in the target capture task requires participants to identify an underlying manifold to effectively map finger joint movements to desired cursor motion towards the targets. To characterize how coordination strategies emerge, we analyzed the simplification of motor-control dimensionality across the three groups by calculating the Variance Accounted For (VAF)~\cite{ranganathan2013learning} by the first few Principal Components (PCs) derived from the hand postures data (finger joint angles $\bs q$). For this analysis, the eight training blocks were condensed into four phases (two blocks per phase) for ease of representation. We also analyze the number of PCs required to explain more than $90\%$ variance in the data across the three groups. We hypothesized that the POMDP group would exhibit a faster increase in top-PC VAF, indicating more efficient discovery of the $2$D task-relevant subspace.

\textbf{Group Comparison:}
As illustrated in Fig.~\ref{fig:synergy_analysis}, the POMDP group achieved higher average VAF in the first four PCs, particularly in later training phases, and required fewer PCs on average to explain $>90\%$ of the variance in joint angle data, \rev{although the results are not statistically significant}. These results provide a structural explanation for the POMDP group's improved task learning and motor performance. By discovering the task space more rapidly than other groups, POMDP group participants successfully coordinated their $20$ finger joints into a simpler, more effective strategy, which is a direct indication of accelerated skill acquisition driven by the optimal nudge feedback.
\begin{table}[b!]
    \centering
    \caption{\textbf{Two-tailed Contrasts for \texttt{RE} Group Differences.} The p-values for the hypothesis that the POMDP group has different \texttt{RE} values across gameplay trials and different number of trials to achieve thresholds compared to the control and the heuristic group are reported. The results show that the POMDP group achieves statistically significantly (at significance level $\alpha=0.10$) lower \texttt{RE} values on average, and takes fewer gameplay trials to achieve \texttt{RE} thresholds early in the training.
    }
    \begin{tabular}{||c|c||c|c|c|c||}
        \hline
        \multicolumn{2}{||c||}{\multirow{2}{*}{\textbf{Metrics}}}&
        \multicolumn{2}{c|}{\textbf{POMDP vs Control}} & \multicolumn{2}{c||}{\textbf{POMDP vs Heuristic}} \\
        \cline{3-6}
        \multicolumn{2}{||c||}{}& \textbf{$\subscr{p}{bonf}$} & \textbf{$\subscr{d}{cohen}$}& \textbf{$\subscr{p}{bonf}$} & \textbf{$\subscr{d}{cohen}$} \\
        \hline \hline
        \multicolumn{2}{||c||}{\textbf{RE Curve}} & \textbf{0.0642} & -0.3863 & \textbf{0.0249} & -0.5899 \\
        \hline
         & 0.14 & {1.0000} & -0.3025 & 1.0000 & -0.0284\\
        \cline{2-6}
        \textbf{RE} & 0.19 & 1.0000 & -0.0400 & 1.0000 & -0.0153\\
        \cline{2-6}
        \textbf{Thresholds} & 0.34 & 0.1705 & -0.8144 & 0.1223 & -0.8932\\
        \cline{2-6}
        & 0.39 & 0.3765 & -0.6116 & 0.4802 & -0.5433\\
        \hline
    \end{tabular}
    \label{table:RE_pvals}
    \vspace{-0.15in}
\end{table}
\subsection{Learning Dynamics and Convergence Analysis}
To quantify the temporal dynamics of motor learning beyond simple performance metrics, we analyzed the evolution of the participants' latent states throughout the experiment. The POMDP framework maintains a continuous belief state $b_k \in \Delta_7$ representing the probability distribution over the IOHMM states at trial $k$.

\textbf{Expected Latent State:}
To visualize the learning trajectory, we computed the \textit{Expected Latent State} for each trial, defined as the probability-weighted average of the state ranks as
\begin{equation}
\mb E[s_k] = \sum_{i=0}^{6} i \cdot b_k(i),
\end{equation}
where $b_k(i)$ is the belief (probability) that the participant is in the $i$-th ranked state at trial $k$. The smaller values of $\mb E[s_k]$ indicate a high probability of having adopted an optimal control strategy.

\begin{figure}
    \centering
    \includegraphics[width=1\linewidth]{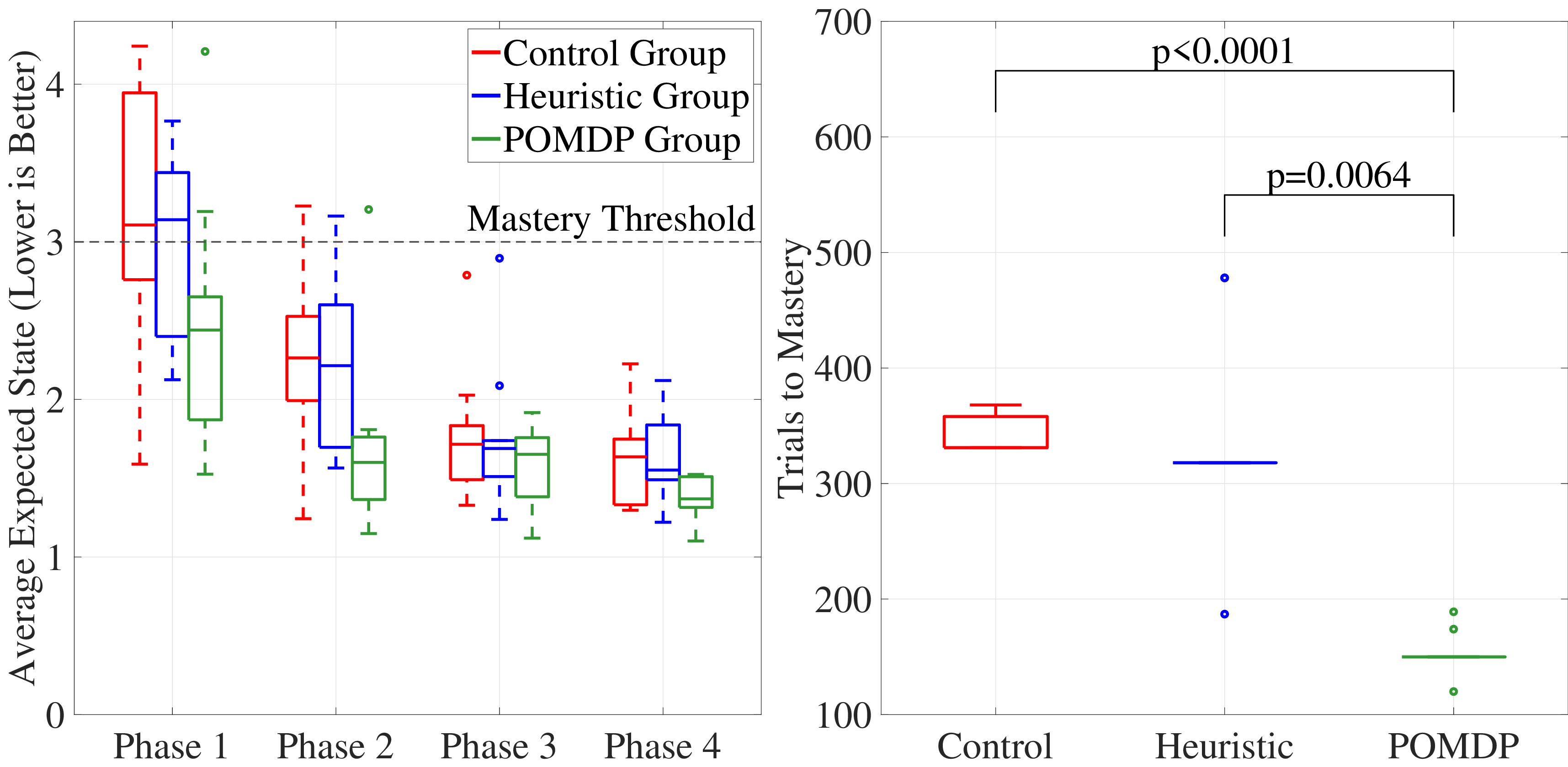}
    \caption{\textbf{Learning Dynamics and Convergence Analysis:} Expected latent state evolution across trials and the number of trials to reach a steady state mastery level ($\mb E[s_k] < 3$) reveals expedited convergence of POMDP group to better performing states and a statistically significantly lower number of trials to mastery, compared to heuristic and control groups.}
    \label{fig:learning}
    \vspace{-0.2in}
\end{figure}

\textbf{Convergence Results:}
\rev{Fig.~\ref{fig:learning} illustrates the distribution of the average expected state for the three experimental groups, aggregated across four phases (two training blocks per phase).
All groups exhibited a characteristic learning progression, initiating Phase $1$ with high expected states, indicating initial exploratory behavior and high motor error. As the experiment progressed, a monotonic decrease in median $\mb E[s_k]$ was observed across all groups, reflecting the acquisition of the BoMI mapping $C$.}

\textbf{Group Comparison:}
\rev{The POMDP group demonstrated a markedly expedited rate of convergence compared to the other two groups. Notably, during Phase $1$, the POMDP group already achieved a median expected state below a pre-defined mastery threshold (defined as $\mb E[s_k] < 3$), whereas the other two groups maintained median states above this threshold. To quantify this temporal advantage, we evaluated the specific number of trials required for participants to reach this steady-state mastery level. The POMDP group reached mastery by trial $150$ (median),} whereas the other two groups reached this threshold statistically significantly later ($p<0.0001$ for Trials(POMDP) $<$ Trials(Control), and $p=0.0064$ for Trials(POMDP) $<$ Trials(Heuristic)).
This suggests that the optimal nudging policy effectively guided participants away from suboptimal local minima and facilitated a more rapid transition into high-performance states. The lower entropy in the belief distribution for the POMDP group towards the end of the session further indicates that these participants not only performed better but had settled into a more stable and confident control strategy compared to other groups.
Since the trials to reach this threshold were not normally distributed across the group participants, we used the Kruskal-Wallis test for statistical significance (at significance level $\alpha<0.05$). This was followed by post-hoc one-tailed contrasts to compare the POMDP group with the other two groups.
\section{Discussion}    \label{sec:discussion}
This work proposes a POMDP-based framework for expediting complex motor skill acquisition and task performance using skill-state-based optimal haptic feedback policy. This proposed approach shows accelerated task performance compared to no feedback and a heuristic feedback design by leveraging the learner's estimated motor skill state. These promising results underscore the significant potential of skill-informed haptic feedback in improving the motor learning process and the corresponding task performance, particularly in novel complex motor skill acquisition tasks.

\subsection{Decoupling Latent Skill from Observable Performance}
A central pillar of our approach is the distinction between task performance and latent skill acquisition. Our results demonstrate that nudge feedback policies targeting simple heuristic policies do not necessarily lead to superior learning and performance outcomes compared to no-feedback baselines (Control Group). This aligns with the guidance hypothesis in motor learning literature~\cite{lee1990role}, which suggests that outcome-based error-correction can engender dependency, allowing the learner to perform well during training without internalizing the task dynamics.

In contrast, the POMDP-based optimal nudging policy unified multiple performance measures into a unified skill state while accounting for the motor noise. By optimizing for long-term cumulative reward, with reward for each trial designed to promote better performance on the current trial as well as all possible task types in the next trial,, the optimal policy likely permitted short-term performance deviations if they facilitated effective exploration of motor space. This supports the notion that effective robotic tutoring should operate as an active teacher, prioritizing the improvement in latent skill state over instantaneous endpoint error corrections.

\subsection{Accelerated Discovery of Better Task Manifolds}
High-dimensional \emph{de-novo} tasks require learners to solve the redundancy problem: identifying a low-dimensional subspace within the high-dimensional task space that maps effectively to the task goal. Our synergy analysis (Fig.~\ref{fig:synergy_analysis}) reveals that the POMDP group exhibited a more rapid consolidation of joint space variance into the first few PCs compared to the other two groups.

This suggests that the skill-state aware nudge feedback did not merely improve task performance, but actively guided participants toward the discovery of the better joint movement subspaces. This behavior is also supported by the learning dynamics and convergence analysis results in Fig.~\ref{fig:learning}. Critically, by penalizing the transition to low-performing skill states, the POMDP policy likely encouraged the utilization of correct fingers sequentially to foster coordinated joint movement strategies toward expedited skill acquisition. This finding is consistent with the Uncontrolled Manifold (UCM) hypothesis, suggesting that optimal assistance facilitates the separation of task-relevant and task-irrelevant variabilities, facilitating the discovery and utilization of effective synergies.

\subsection{Interpretability of Data-driven Behavior Models}
While data-driven models are often criticized for their lack of interpretability (black-box models) compared to mechanistic first-principle models, our IOHMM successfully captured biomechanically relevant features of the task. The state transition matrices in Fig.~\ref{fig:STM} revealed that the haptic nudges to the thumb (finger index $1$) significantly increased the probability of transitioning to higher skill states, whereas nudges to the pinky (finger index $5$) had a smaller impact. Furthermore, the STMs for neighboring nudged fingers are closer to each other than the rest.

This aligns with the kinematic realities of our setup and the BoMI, where the thumb possesses a larger range of motion and greater contribution to cursor motions than the other fingers. The model learned these biomechanical constraints solely from the input-output data, without explicit biomechanical modeling. This validates the IOHMM not just as a control tool, but as a diagnostic tool capable of identifying which fingers are most receptive to haptic interventions at specific stages of learning. Such interpretability is crucial for clinical application, where therapists must trust the agent to prescribe physiologically valid movement patterns.

\subsection{Limitations and Future Directions}
While our results are promising, several limitations warrant discussion. First, the discretization of skill state into $N=7$ states is an abstraction of the continuous motor learning process. While necessary for the tractability of the QMDP solver, this quantization may mask subtle changes/adjustments in a strategy that occur within a single state. Future work could explore continuous state POMDP solvers or hybrid approaches to capture the finer details/granularities in the learning dynamics. However, data constraints may make fitting a continuous state model reliably a challenging task.

Second, our study validated the proposed framework on healthy humans learning a novel motor task. While this paradigm is an established proxy for the motor re-learning process, post-neurological injury populations introduce state-dependent constraints owing to specific impairments (eg, spasticity, paresis) that are not modeled in healthy participants. However, the data-driven nature of our IOHMM framework is agnostic to specific learning dynamics. In principle, the same approach could be applied to patient data to learn the personalized transition model; however, this would necessitate the use of individualized models instead of a population-average model. Another possibility is to estimate a population average model, personalizing it to each individual using real-time input-output data during the experiment. Future efforts could focus on transitioning this framework to clinical cohorts to validate its efficacy in neuro-rehabilitation.


\bibliographystyle{IEEEtran}

\bibliography{masterbib, mybib}

\end{document}